\documentclass[12pt,a4paper,onecolumn,journal,draftclsnofoot, 12pt]{IEEEtran}
\usepackage[LGR,T1]{fontenc}
\usepackage[latin9]{inputenc}
\usepackage{color}
\usepackage{array}
\usepackage{booktabs}
\usepackage{textcomp}
\usepackage{mathtools}
\usepackage{multirow}
\usepackage{amsmath}
\usepackage{amssymb}
\usepackage{graphicx}

\makeatletter

\pdfpageheight\paperheight
\pdfpagewidth\paperwidth

\providecommand{\tabularnewline}{\\}

\usepackage{array}
\usepackage{units}
\usepackage{multirow}
\usepackage{xcolor}

\providecommand{\tabularnewline}{\\}

\makeatother

\begin{document}
\title{An Energy and Carbon Footprint Analysis of Distributed and Federated
Learning}
\author{Stefano Savazzi, Vittorio Rampa, Sanaz Kianoush, Mehdi Bennis}

\maketitle

\begin{abstract}
Classical and centralized Artificial Intelligence (AI) methods require
moving data from producers (sensors, machines) to energy hungry data
centers, raising environmental concerns due to computational and communication
resource demands, while violating privacy. Emerging alternatives to
mitigate such high energy costs propose to efficiently distribute,
or federate, the learning tasks across devices, which are typically
low-power. This paper proposes a novel framework for the analysis
of energy and carbon footprints in distributed and federated learning
(FL). The proposed framework quantifies both the energy footprints
and the carbon equivalent emissions for vanilla FL methods and consensus-based
fully decentralized approaches. We discuss optimal bounds and operational
points that support green FL designs and underpin their sustainability
assessment. Two case studies from emerging 5G industry verticals are
analyzed: these quantify the environmental footprints of continual
and reinforcement learning setups, where the training process is repeated
periodically for continuous improvements. For all cases, sustainability
of distributed learning relies on the fulfillment of specific requirements
on communication efficiency and learner population size. Energy and
test accuracy should be also traded off considering the model and
the data footprints for the targeted industrial applications. 
\end{abstract}

\IEEEpeerreviewmaketitle{}

\section{Introduction}

Training deep Machine Learning (ML) models at the network edge has
reached notable gains in terms of accuracy across many tasks, applications
and scenarios. However, such improvements have been acquired at the
cost of large computational and communication resources, as well as
significant energy footprints which are currently overlooked. Vanilla
ML requires all training procedures to be conducted inside data centers
\cite{kone1} that collect data from producers, such as sensors, machines
and personal devices. Centralized based learning requires high energy
costs for networking and data maintenance, while introduces privacy
issues as well. In addition, many mission critical ML tasks require
data centers to learn continuously \cite{commmag} to track changes
in dataset distributions. Data centers are thus becoming more and
more energy-hungry and responsible of significant CO2 (carbon) emissions.
These amount to about $15$\% of the equivalent global Green House
Gas (GHG) emissions of the entire Information and Communication Technology
(ICT) ecosystem \cite{first_look}, estimated as $0.5\div1$ GtCO2e
(gigatonnes of CO2 equivalent emissions) in 2020. Electricity consumed
by global data centers is also estimated to be between $1.1$\% and
$1.5$\% of the total worldwide electricity use.

Emerging alternatives to centralized big-data analytics enable decisions
to be made at much more granular levels \cite{drl}. The Federated
Learning (FL) approach \cite{key-7} is a recently proposed distributed
AI paradigm \cite{macmahan} in line with this trend. Under FL, the
ML model parameters, e.g. the weights and biases $\mathbf{W}$ of
Deep Neural Networks (DNN), are collectively optimized across several
resource-constrained edge/fog devices, that act as \emph{both} data
producers \emph{and} local learners. FL distributes the computing
tasks across many devices characterized by low-power consumption profiles,
compared with data centers, and each device has access to small datasets
\cite{commmag} only. With a judicious system design taking into account
both accuracy and energy, FL is expected to bring significant reduction
in terms of energy footprints, obviating the need for a large centralized
infrastructure for cooling or power delivery. 

\begin{figure}[!t]
\centering \includegraphics[scale=0.5]{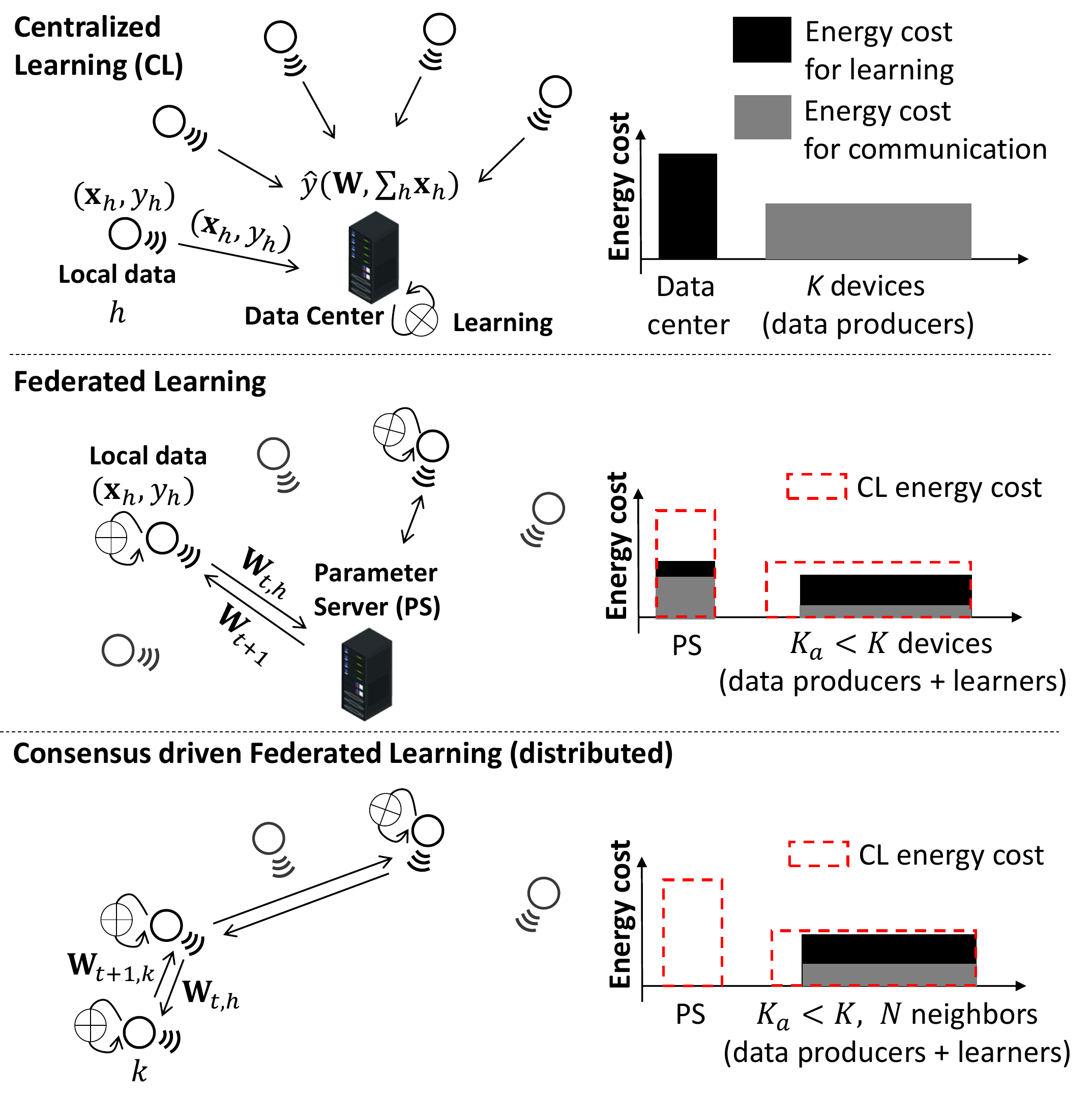} 
 \protect\caption{\label{intro} From top to bottom: Centralized Learning (CL), Federated
Learning (FL) coordinated by the Parameter Server (PS), namely Federated
Averaging (FA), and Consensus-driven Federated Learning, enabling
fully decentralized implementations (i.e., without PS).}
\vspace{-0.5cm}
\end{figure}

\subsection{Federated, consensus-driven and centralized learning: related works}

As shown in Fig. \ref{intro}, vanilla FL algorithms, such as Federated
Averaging (FA) \cite{drl}, allow devices to learn a local model under
the orchestration of a Parameter Server (PS). Devices might be either
co-located in the same geographic area or distributed in different
zones. Typically, the PS aggregates the received local models to obtain
a \emph{global model} that is fed back to the devices. The PS functions
are substantially less energy-hungry compared to centralized learning
(CL) and can be implemented at the network edge. Different FL implementations
have emerged in the past few years \cite{survey_FL_iot} targeting
several scenarios \cite{commmag,resource_efficency}. The optimization
of the population of learners is critical in FL and is typically explored
when power and bandwidth of devices are limited \cite{jointopt}.
Delay and energy tradeoffs between learning and communication are
considered in \cite{harvesting}, while the problem of transmission
scheduling in small-scale fading is analyzed in \cite{energy2}. Quantization
and compression of model parameters, or gradients, is useful to minimize
bandwidth usage and to avoid straggler effects \cite{fedpaq,GADMM,compression}.
For a comprehensive survey on FL please refer to \cite{drl}.

The above mentioned designs leverage on a server-client architecture
\cite{survey_FL_iot} where the PS represents a single-point of failure
and penalizes scalability. In analogy with distributed ledger, consensus-driven
learning replaces the PS functions as it lets the local model parameters
be consensually shared and synchronized across multiple devices via
device-to-device (D2D) or sidelink communications \cite{commmag}.
Such fully decentralized solution relies \emph{solely} on in-network
processing and a consensus-based federation model, i.e., often based
on distributed weighted averaging \cite{gossip-1}-\cite{chen}. 

Regardless of whether the network architecture is server-client or
decentralized via consensus methods, distributed learning requires
many communication rounds for convergence that often consist of UpLink
(UL) and DownLink (DL) communications through a Wireless Wide Area
access and core Network (WWAN). Large energy bills might be therefore
observed for communication \cite{harvesting}. The choice between
distributed and centralized learning targeting sustainable, energy-aware
designs is thus expected to be driven by communication vs computing
cost balancing.

\subsection{Contributions}

The paper develops a novel framework for the analysis of energy and
carbon footprints in distributed ML, including, for the first time,
comparisons and tradeoff considerations about vanilla, consensus-driven
FL and centralized learning on the data center. Despite initial attempts
to assess FL resource efficiency \cite{resource_efficency}, optimize
delay/energy \cite{harvesting,energy2} and carbon footprints \cite{first_look,pimrc},
quantifying the optimal operating points towards green or sustainable
setups is yet unexplored. To fill this void, the paper dives into
the implementations of selected distributed learning tools, and discusses
optimized designs and operational conditions with respect to energy-efficiency
and low carbon emissions. A general framework is first developed for
energy and carbon footprint assessment: differently from \cite{first_look,harvesting},
it considers all the system parameters and components of the learning
process, including the PS, required in vanilla FL, as well as the
distributed model averaging, required in decentralized setups. Next,
sustainable regions in the parameter space are identified to steer
the choice between centralized and distributed training in practical
setups: the regions highlight bounds, or necessary conditions on communication,
computing efficiencies, data and model size that make FL more promising
than CL with respect to its carbon footprint. Unlike classical FL
platforms \cite{drl,harvesting} defined on top of edge devices with
large computing power \cite{resource_efficency}, the paper focuses
on learning tasks suitable for low-power embedded wireless devices
(e.g., robots, vehicles, drones) typically adopted in Industrial Internet
of Things (IIoT) applications \cite{survey_FL_iot}. These setups
are characterized by devices/learners retaining small training datasets,
while running medium/small-sized ML models due to their constrained
internal memory. Devices are also equipped with low-power radio interfaces
supporting deep sleep modes, cellular (e.g., 5G and NB-IoT), as well
as direct mode, or sidelink, communication interfaces.

The proposed co-design of learning and communication is validated
using real world data and a test-bed platform characterized by low-power
devices implementing real-time distributed model training on top of
the Message Queuing Telemetry Transport (MQTT) protocol. The real-time
test-bed is used to measure the training time for different learning
policies, while a \emph{what-if} analysis quantifies the estimated
energy and carbon footprints considering the impact of various communication
settings, as well as the influence of the learner population size. 

\subsection{Paper structure and organization}

The paper is organized as follows. Section \ref{sec:Energy-footprint-modeling}
describes the framework for energy consumption evaluation of different
distributed learning strategies. Energy modelling is carried out separately
for centralized learning (Sect.\ref{subsec:Centralized-Learning}),
vanilla FA (Sect. \ref{subsec:Federated-Averaging}), including the
novel policy FA-D that keeps inactive learners in deep-sleep, and
decentralized consensus-driven FA (Sect. \ref{subsec:Consensus-driven-Federated-Avera}).
Necessary adaptations of the proposed energy models to \emph{continual}
and \emph{Reinforcement Learning} (RL) paradigms \cite{rl} are also
discussed in Sect. \ref{subsec:Adaptations-to-continual,}. 

Sect. \ref{sec:Carbon-footprint-assessment} considers the carbon
footprints (Sect. \ref{subsec:Energy-efficiency-and}) and the sustainable
regions (Sect. \ref{subsec:Carbon-aware-design-constraints}) that
provide necessary requirements to steer the choice between centralized
and distributed learning paradigms. Requirements are verified first
in Sect. \ref{subsec:Examples-with-MNIST} using the MNIST \cite{mnist}
and the CIFAR10 \cite{cifar10} datasets (Sect. \ref{subsec:Impact-of-conditions})
separately and through comparative analysis (Sect. \ref{subsec:MNIST-and-CIFAR:}).
Next, practical design problems are explored in Sect. \ref{sec:A-case-study}
targeting two industry relevant use cases. In particular, the FL network
implementation is based on the MQTT protocol (Sect. \ref{subsec:Networking-and-MQTT})
while 4 communication efficiency profiles are considered (Sect. \ref{subsec:Communication-efficiencies}).
Both continual (Sect. \ref{subsec:Continual-and-few-shot}) and RL
(Sect. \ref{subsec:Reinforcement-learning}) scenarios are analyzed
where the training process must be periodically repeated using new
input data that change frequently to follow either a time-varying
industrial process or the same model under training. These tools are
critical in 5G industry verticals, i.e., Industry 4.0 (I4.0) applications,
and could raise significant energy concerns unless specific actions
are taken. Finally, conclusion are summarized in Sect. \ref{sec:Conclusions-and-open}.

\section{Energy footprint modeling framework \label{sec:Energy-footprint-modeling}}

The proposed framework provides insights into how the different components
of the CL (i.e., the datacenter) and FL architectures (i.e. the local
learners, the core network and the PS), contribute to the energy bill
and to the carbon emissions in terms of system accuracy and number
of rounds. The learning system consists of $K$ devices and one data
center ($k=0$). Each device $k>0$ has a dataset $\mathcal{E}_{k}$
of examples $(\mathbf{x}_{h},y_{h})$ that are typically collected
independently. In supervised learning, examples $\mathbf{x}_{h}$
are labelled as $y_{h}$ for training. Unsupervised setups, i.e.,
RL, are discussed in Sect. \ref{subsec:Adaptations-to-continual,}.
For all cases, the objective of the learning system is to train a
DNN model $\hat{y}(\mathbf{W};\mathbf{x})$ that transforms the input
data $\mathbf{x}$ into the desired $C$ outputs $\hat{y}\in\left\{ y_{c}\right\} _{c=1}^{C}$,
i.e., the output classes. Model parameters are here specified by the
matrix $\mathbf{W}$ \cite{drl}. The training system uses the examples
in $\mathcal{E}=\bigcup_{k=1}^{K}\mathcal{E}_{k}$ to minimize a loss
function $L$ of the form
\begin{equation}
L(\mathcal{E}|\mathbf{W})=\sum_{k=1}^{K}L_{k}(\mathcal{E}_{k}|\mathbf{W}).\label{eq:loss}
\end{equation}
In what follows, unless stated otherwise, we use the cross-entropy
loss function \cite{drl}, namely $L_{k}=-\sum_{h}y_{h}\log\left[\hat{y}(\mathbf{W};\mathbf{x}_{h})\right]$.
Minimization of (\ref{eq:loss}) is typically iterative and gradient-based
\cite{key-7}: it runs over a pre-defined number ($n$) of learning
rounds that depend on a specified target loss threshold, namely $L(\mathcal{E}|\mathbf{W})\leq\overline{\xi}$,
i.e., corresponding to a required accuracy. For example, using Stochastic
Gradient Descent (SGD)

\begin{equation}
\mathbf{W}_{t+1}\leftarrow\mathbf{W}_{t}-\mu_{s}\times\nabla L_{k}(\mathcal{E}_{k}|\mathbf{W}_{t}),\label{nofederation}
\end{equation}
examples $\mathcal{E}_{k}$ are drawn randomly from the full training
set $\mathcal{E}$, while $\mu_{s}$ is the step-size and $\nabla L_{k}(\mathcal{E}_{k}|\mathbf{W}_{t})=\nabla_{\mathbf{W}_{t}}\left[L(\mathcal{E}_{k}|\mathbf{W}_{t})\right]$
is the gradient of the loss function (\ref{eq:loss}) over the assigned
batches of data $\mathcal{E}_{k}$ given the model $\mathbf{W}_{t}$.

The energy footprint of device $k$, namely the total amount of energy
consumed by the learning process, is broken down into computing and
communication components. Considering one learning round, all the
energy costs are modelled as a function of the computing energy $E_{k}^{(\mathrm{C})}$
required by the optimizer (\ref{nofederation}) and the energy $E_{k,h}^{(\mathrm{T})}$
per correctly received/transmitted bit over the wireless link ($k,h$).
The training data or the DNN model parameters are quantized before
transmission into $b(\mathcal{E}_{k})$ and \textbf{$b(\mathbf{W})$}
bits, respectively. The quantization scheme assigns a fixed number
of bits (here 32 bits) to each parameter of the DNN model. Compression
techniques such as model pruning, sparsification \cite{comp}, parameter
selection and/or differential transmission schemes \cite{fedpaq,camad}
are extremely helpful to scale down the footprint \textbf{$b(\mathbf{W})$}
in large DNNs. However, due to the great variety of compression techniques,
we have considered here only a simple quantization scheme. The modifications
to include a particular technique can be trivially detailed using
the specific data $b(\mathcal{E}_{k})$ and model size \textbf{$b(\mathbf{W})$}
as the result of the compression processes.

We define the energy cost for UL communication with the data center
(or the PS), co-located with the Base Station (BS), as $E_{k,0}^{(\mathrm{T})}$.
Similarly, $E_{0,k}^{(\mathrm{T})}$ refers to DL communication, i.e.,
from the PS, or the BS, to the \emph{k}-th device. Communication costs
incorporate both transmission and decoding operations \cite{survey_energy_5G}.
In contrast to digital implementations, analog FL designs \cite{jointopt,digvanalog}
get over the restrictions of time scheduled access: rather than spending
energy $E_{k,h}^{(\mathrm{T})}$ for sending bits over orthogonal
links, they exploit the superposition property of wireless transmissions
for analog aggregation \cite{analog2}, scaling down the number of
channel uses, and the communication cost. Although not considered
in this paper, comparing analog and digital FL in the light of energy
footprints is an open problem of wide interest. The energy $E_{k}^{(\mathrm{C})}$
for computing includes the cost of the learning round, namely the
local gradient-based optimizer and the data storage costs. All costs
are quantified on average: notice that routing through the radio access
and the core network can vary but might be assumed as stationary apart
from failures or replacements.

\subsection{Centralized Learning\label{subsec:Centralized-Learning}}

Under CL, the model training is carried out inside the data center
$k=0$, assumed co-located with the BS, and exploits the processing
power provided by racks of CPU (Central Processing Units), GPU (Graphic
Processing Units) and other specialized AI accelerators such as NPU
(Neural Processing Units) and TPU (Tensor Processing Units). Therefore,
the energy cost per round $E_{0}^{(\mathrm{C})}=P_{0}\cdot T_{0}\cdot B$
depends on the power consumption $P_{0}$ \cite{pimrc} of the aforementioned
hardware units, the time span $T_{0}$ required for processing an
individual batch of data, i.e. minimizing the loss $L(\cdot|\mathbf{W})$,
and the number $B$ of batches per round \cite{harvesting}. Data
are collected independently by the devices while we neglect here the
cost of initial dataset loading since it is a one-step process only.
For $n$ rounds, chosen to satisfy a target loss $\overline{\xi}$,
the total energy $E_{\mathrm{CL}}$ in Joule {[}J{]} consists of computing
$E_{\mathrm{CL}}^{(\mathrm{L})}$ and communication $E_{\mathrm{CL}}^{(\mathrm{C})}$
costs:

\begin{equation}
E_{\mathrm{CL}}(n)=\underset{E_{\mathrm{CL}}^{(\mathrm{L})}}{\underbrace{\gamma\cdot n\cdot E_{0}^{(\mathrm{C})}}}+\underset{E_{\mathrm{CL}}^{(\mathrm{C})}}{\underbrace{\alpha\cdot\sum_{k=1}^{K}b(\mathcal{E}_{k})\cdot E_{k,0}^{(\mathrm{T})}}.}\label{eq:cl}
\end{equation}
The energy cost for computing inside the data center scales as $E_{\mathrm{CL}}^{(\mathrm{L})}(n)=\gamma\cdot n\cdot E_{0}^{(\mathrm{C})}$
with\textbf{ $\mathrm{\gamma}>1$} being the Power Usage Effectiveness
(PUE) of the data center \cite{data_center,cooling}. The PUE accounts
for the additional power consumed for data storage, power delivery
and cooling: $\gamma$ values are in the range $1.1\div1.8$ \cite{cooling,trend-1}.
The energy consumed for communication of raw data is $E_{\mathrm{CL}}^{(\mathrm{C})}=\alpha\sum_{k=1}^{K}b(\mathcal{E}_{k})\cdot E_{k,0}^{(\mathrm{T})}$
and quantifies the energy for moving $b(\mathcal{E}_{k})$ bits of
the $k$-th local training dataset $\mathcal{E}_{k}$ for $\alpha$
times and for all the $K$ devices. Parameter $\alpha$ counts the
number of times a new training dataset is uploaded by the devices
while learning is in progress. For example, in one-time learning processes,
the full training dataset is available by devices before the training
starts: therefore, it is $\alpha=1$. On the other hand, in continual
learning applications, devices produce new training data continuously
\cite{commmag,rl}: data thus needs to be moved while learning is
in progress, therefore it is $1<\alpha\leq n$.

\begin{figure}[!t]
\centering \includegraphics[scale=0.5]{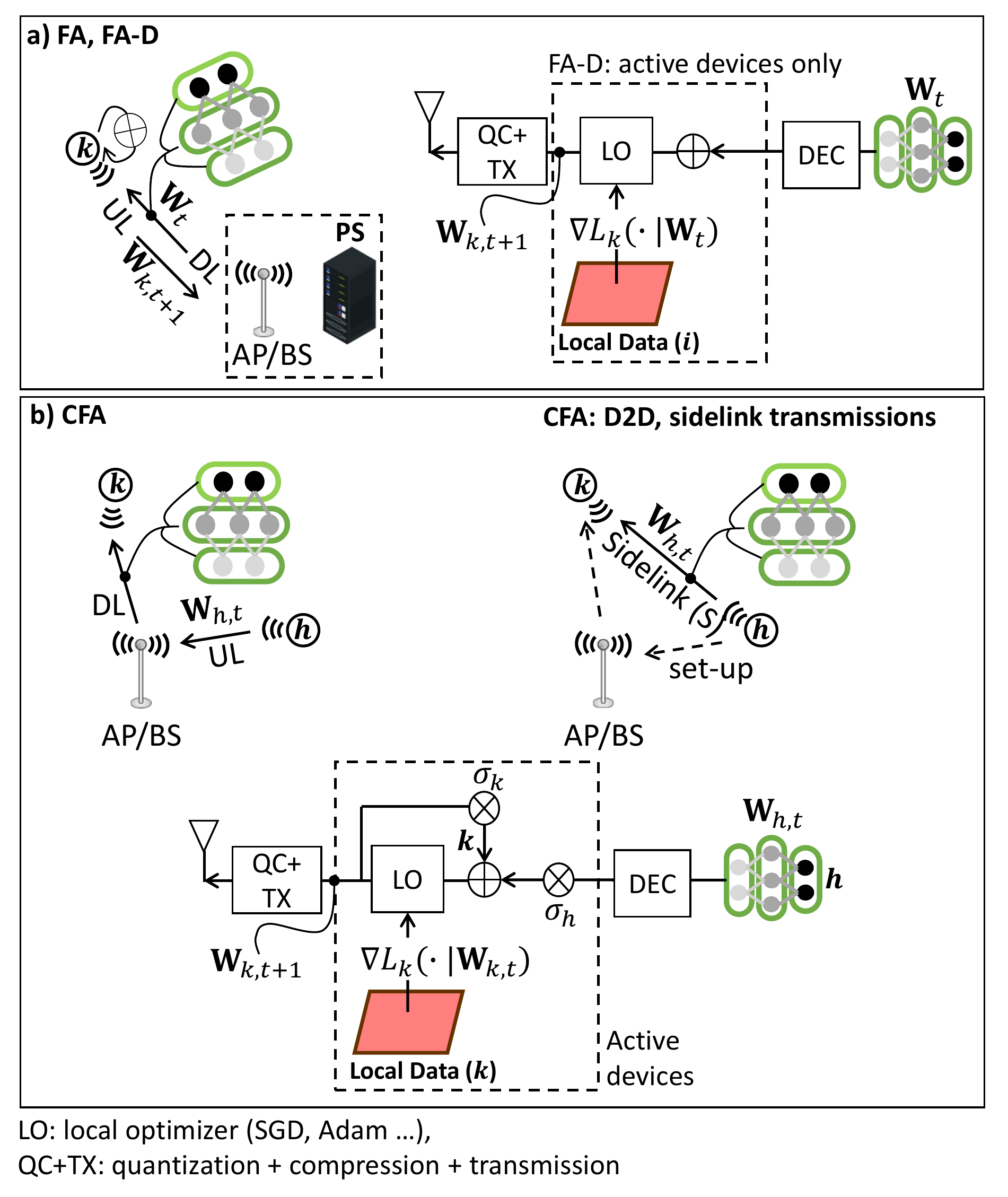} 
 \protect\caption{\label{FA-CFA-CFAGE} Distributed learning algorithms: a) Federated
Averaging (FA, FA-D), b) CFA with sidelink communications.}
\vspace{-0.5cm}
\end{figure}

\subsection{Federated learning with server (FA) and deep-sleep (FA-D)\label{subsec:Federated-Averaging}}

Unlike CL, FL distributes the learning process across a selected,
typically time-varying \cite{drl}, subset $\mathcal{N}_{t}$ of $K_{a}<K$
\emph{active} devices as shown in Fig. \ref{intro}(a). At each round
$t$, the local dataset $\mathcal{E}_{k}$ is used to train the local
model $\mathbf{W}_{k,t}$ using a gradient-based Local Optimizer (LO)
and a global model instance ($\mathbf{W}_{t}$), obtained from the
server (PS). The goal is to minimize the local loss $L_{k}$ as 
\begin{equation}
\mathbf{W}_{k,t}=\underset{\mathbf{W}}{\mathrm{arg}\mathrm{min}}\thinspace L_{k}(\cdot|\mathbf{W})+\upsilon\left\Vert \mathbf{W}-\mathbf{W}_{t}\right\Vert ^{2},\label{eq:LO}
\end{equation}
where $\upsilon\left\Vert \mathbf{W}-\mathbf{W}_{t}\right\Vert ^{2}$
is the proximal term \cite{energy2,fedprox} often used when data
is non-identically distributed (non-IID). The local model $\mathbf{W}_{k,t}$
is then forwarded to the PS \cite{drl} over the UL. The PS is co-located
with the BS and is in charge of updating the global model $\mathbf{W}_{t+1}$
for the following round $t+1$ through the aggregation of the $K_{a}$
received models \cite{key-7}: 
\begin{equation}
\mathbf{W}_{t+1}\leftarrow\sum_{k\in\mathcal{N}_{t}}\sigma_{k}\cdot\mathbf{W}_{k,t},\label{eq:farho}
\end{equation}
with $\sigma_{k}=\tfrac{Q_{k}}{Q}$ being the ratio between the number
of local examples $Q_{k}$ and global $Q=\sum_{k\in\mathcal{N}_{t}}Q_{k}$
ones, respectively. Notice that for IID data distributions, it is
$\sigma_{k}=\tfrac{1}{K_{a}}$ \cite{macmahan}. The new model $\mathbf{W}_{t+1}$
is finally sent back to the devices over the DL. 

\textbf{Federated Averaging (FA)}. In vanilla FL methods, such as
FA, the PS chooses $K_{a}<K$ active devices on each round to communicate
their local model. During this process, the remaining $K-K_{a}$ devices
are powered on \cite{key-7,macmahan} as they run the LO (\ref{eq:LO})
and decode the updated global model $\mathbf{W}_{t}$ obtained from
the PS. For $n$ rounds, the total end-to-end energy can be written
as 
\begin{equation}
E_{\mathrm{FA}}(n)=E_{\mathrm{FA}}^{(\mathrm{L})}(n)+E_{\mathrm{FA}}^{(\mathrm{C})}(n),\label{eq:fagen}
\end{equation}
namely, the superposition of the computing $E_{\mathrm{FA}}^{(\mathrm{L})}(n)$
and the communication $E_{\mathrm{FA}}^{(\mathrm{C})}(n)$ costs,
considering both the devices and the PS consumption. In particular:

\begin{equation}
\begin{array}{c}
E_{\mathrm{FA}}^{(\mathrm{L})}(n)=\gamma\cdot n\cdot\beta\cdot E_{0}^{(\mathrm{C})}\,+n\cdot\sum_{k=1}^{K}E_{k}^{(\mathrm{C})},\\
E_{\mathrm{FA}}^{(\mathrm{C})}(n)=b(\mathbf{W})\left[n\cdot\sum_{k=1}^{K}\gamma\cdot E_{0,k}^{(\mathrm{T})}+\sum_{t=1}^{n}\sum_{k\in\mathcal{N}_{t}}E_{k,0}^{(\mathrm{T})}\right].
\end{array}\begin{aligned}\end{aligned}
\label{eq:fa}
\end{equation}
The energy cost for computing $E_{\mathrm{FA}}^{(\mathrm{L})}(n)$
in (\ref{eq:fa}) is due to the PS energy $\gamma\cdot n\cdot\beta\cdot E_{0}^{(\mathrm{C})}$
needed for model averaging and the LO, $n\cdot\sum_{k=1}^{K}E_{k}^{(\mathrm{C})}$
that is implemented by all the $K$ devices \cite{harvesting}. The
energy for model averaging $\beta\cdot E_{0}^{(\mathrm{C})}$ is considerably
smaller than the gradient-based optimization term $E_{0}^{(\mathrm{C})}$
on the data center, therefore it is $\beta\ll1$. In IoT setups, the
devices are also usually equipped with embedded low-consumption CPUs,
SoCs (System on Chips) or $\mu$CU (microcontrollers): thus, it is
reasonable to assume $E_{k}^{(\mathrm{C})}<E_{0}^{(\mathrm{C})}$.
The communication energy $E_{\mathrm{FA}}^{(\mathrm{C})}(n)$ in (\ref{eq:fa})
models the cost of the global model transmission over the DL, i.e.,
$n\cdot b(\mathbf{W})\sum_{k=1}^{K}\gamma\cdot E_{0,k}^{(\mathrm{T})}$,
and the UL communication from selected $K_{a}$ active devices, i.e.,
$b(\mathbf{W})\sum_{t=1}^{n}\sum_{k\in\mathcal{N}_{t}}E_{k,0}^{(\mathrm{T})}$
in the set $\mathcal{N}_{t}$: as depicted in Fig. \ref{FA-CFA-CFAGE}(a),
it is assumed that the PS is co-located with the BS. The model size
$b(\mathbf{W})$ quantifies the size in bits of model parameters to
be exchanged\footnote{More precisely, $b(\mathbf{W})$ quantifies the size of the subset
of the (trainable) model layers, or parameters, exchanged on each
FL round: however, to simplify the reasoning, it is herein referred
to as ``model size''. }. As analyzed in the Sect. \ref{sec:Carbon-footprint-assessment},
both model $b(\mathbf{W})$ and local data size $b(\mathcal{E}_{k})$
are critical for sustainable designs. Notice that, as opposed to data,
$b(\mathbf{W})$ is roughly the same for each device, although small
changes might be observed when using lossy compression, sparsification
or parameter selection methods \cite{fedpaq,comp,camad}.

\textbf{Federated Averaging with deep-sleep (FA-D)}. An alternative
to FA, referred to as FA-D, lets the inactive $K-K_{a}$ devices to
turn off their computing hardware and communication interface, and
set into a deep-sleep mode, when not needed by the PS. Deep sleep
mode should be paired with efficient hardware able to power up and
down in fractions of milliseconds: this is also a key component of
5G \cite{survey_energy_5G} and supported by popular interfaces, such
as NB-IoT \cite{nbiot}. In contrast to vanilla FA, FA-D limits the
consumption to the $K_{a}$ active devices, while these might change
from round to round on a duty-cycle basis. As verified in the following,
FA-D brings significant per-round energy benefits, in exchange for
larger number of rounds $n_{\mathrm{FA-D}}>n_{\mathrm{FA}}$, for
the same target accuracy $\overline{\xi}$. Considering the duty cycling
process, the energy footprint of FA-D becomes 
\begin{equation}
E_{\mathrm{FA-D}}(n)=E_{\mathrm{FA-D}}^{(\mathrm{L})}(n)+E_{\mathrm{FA-D}}^{(\mathrm{C})}(n),\label{eq:fad}
\end{equation}
now with
\begin{equation}
\begin{array}{c}
E_{\mathrm{FA-D}}^{(\mathrm{L})}(n)=\gamma\cdot n\cdot\beta\cdot E_{0}^{(\mathrm{C})}\,+\sum_{t=1}^{n}\left[\sum_{k\in\mathcal{N}_{t}}E_{k}^{(\mathrm{C})}+\sum_{k\notin\mathcal{N}_{t}}E_{k,\mathrm{sleep}}^{(\mathrm{C})}\right],\\
E_{\mathrm{FA-D}}^{(\mathrm{C})}(n)=b(\mathbf{W})\cdot\sum_{t=1}^{n}\sum_{k\in\mathcal{N}_{t}}\left[\gamma\cdot E_{0,k}^{(\mathrm{T})}+E_{k,0}^{(\mathrm{T})}\right].
\end{array}\begin{aligned}\end{aligned}
\label{eq:fa-d}
\end{equation}
Notice that the LO cost is now constrained to the active devices,
$\sum_{t=1}^{n}\sum_{k\in\mathcal{N}_{t}}E_{k}^{(\mathrm{C})}$ as
with global model communication, $b(\mathbf{W})\cdot\sum_{t=1}^{n}\sum_{k\in\mathcal{N}_{t}}\gamma\cdot E_{0,k}^{(\mathrm{T})}$.
Finally, $E_{k,\mathrm{sleep}}^{(\mathrm{C})}$ accounts for the energy
consumption of inactive devices in deep-sleep mode (if not negligible
w.r.t. $E_{k}^{(\mathrm{C})}$ ).

\subsection{Consensus-driven Federated Averaging (CFA)\label{subsec:Consensus-driven-Federated-Avera}}

Decentralized FL techniques \cite{cfa,digvanalog,chen} let the devices
mutually exchange their local models, possibly over peer-to-peer networks.
Unlike with FA, the PS is not needed as it is replaced by a consensus
mechanism. In particular, we propose a consensus-driven FA approach
(CFA) where the federated nodes exchange the local ML model parameters
and update them sequentially by distributed averaging. As shown in
Fig. \ref{FA-CFA-CFAGE}(b), devices mutually exchange their local
model parameters $\mathbf{W}_{k,t}$ with an assigned number $N$
of neighbors \cite{fedpaq,GADMM,digvanalog}. Distributed weighted
averaging \cite{cfa,chen} is used to combine the received models. 

Let $\mathcal{N}_{k,t}$ be the set that contains the $N$ chosen
neighbors of node $k$ at round $t$: on every new round ($t>0$),
the device updates the local model $\mathbf{W}_{k,t}$ using the parameters
$\mathbf{W}_{h,t}$ obtained from the neighbor device(s) as 
\begin{equation}
\mathbf{W}_{k,t+1}\leftarrow\mathbf{W}_{k,t}+\sum_{h\in\mathcal{N}_{k,t}}\sigma_{h,k}\cdot(\mathbf{W}_{h,t}-\mathbf{W}_{k,t}),\label{eq:averaging}
\end{equation}
where the weights $\sigma_{h,k}$ are chosen to guarantee a stable
solution as
\begin{equation}
\sigma_{h,k}=\frac{Q_{h}}{\sum_{h\in\mathcal{N}_{k,t}}Q_{h}},\label{eq:weights}
\end{equation}
and for IID data, it is $\sigma_{h,k}=\tfrac{1}{N}$, $\forall h,k$.
Distributed weighted averaging (\ref{eq:averaging}) is followed by
LO (\ref{eq:LO}) on $\mathcal{E}_{k}$ that can also include proximal
regularization \cite{fedprox}. For $K_{a}<K$ active devices in the
set $\mathcal{N}_{t}$ and $n$ rounds, the energy footprint is again
broken down into learning and communication costs as 
\begin{equation}
E_{\mathrm{CFA}}(n)=E_{\mathrm{CFA}}^{(\mathrm{L})}(n)+E_{\mathrm{CFA}}^{(\mathrm{C})}(n),\label{eq:cfa_general}
\end{equation}
 with

\begin{equation}
\begin{array}{c}
E_{\mathrm{CFA}}^{(\mathrm{L})}(n)=\sum_{t=1}^{n}\left[\sum_{k\in\mathcal{N}_{t}}E_{k}^{(\mathrm{C})}+\sum_{k\notin\mathcal{N}_{t}}E_{k,\mathrm{sleep}}^{(\mathrm{C})}\right],\\
E_{\mathrm{CFA}}^{(\mathrm{C})}(n)=\sum_{t=1}^{n}\sum_{k\in\mathcal{N}_{t}}\sum_{h\in\mathcal{N}_{k,t}}b(\mathbf{W}_{k,t})\cdot E_{k,h}^{(\mathrm{T})}.
\end{array}\begin{aligned}\end{aligned}
\label{eq:cfa}
\end{equation}
The sum $\sum_{h\in\mathcal{N}_{k,t}}b(\mathbf{W}_{k,t})\cdot E_{k,h}^{(\mathrm{T})}$
now models the total energy spent by the device $k$ to diffuse the
local model parameters to selected neighbors $h$ in the set $\mathcal{N}_{t}$
at round $t$. Since the PS is not used, CFA is particularly promising
in reducing the energy footprint. 

CFA requires sending data over peer-to-peer links ($k,h$): in WWAN,
each link is implemented by UL transmission from the source $k$ to
the core network access point (i.e., router or BS), followed by a
DL communication from the router(s) to the destination device $h$,
therefore 
\begin{equation}
E_{k,h}^{(\mathrm{T})}=E_{k,0}^{(\mathrm{T})}+\gamma\cdot E_{0,h}^{(\mathrm{T})},\label{eq:cellularsidelink}
\end{equation}
where $\gamma$ is the PUE of the BS or router hardware (if any).
Alternatively, CFA can leverage direct, or sidelink, transmissions
\cite{d2dcomm}. Sidelink (SL) features were introduced by 3GPP (PC5
interface) in release $12$ and $13$, namely the proximity service
(ProSe). They support direct (D2D) communications with minor involvement
of the BS, or eNB \cite{6g,6ggreen}. Since the signal relay through
the BS is not needed, excluding an initial link setup for resource
allocation, sidelink communications are expected to make a significant
step towards green and sustainable networks, serving as enabling technology
for an efficient implementation of CFA. Both communication architectures
will be addressed next.

\subsection{Continual and reinforcement learning paradigms \label{subsec:Adaptations-to-continual,}}

The energy framework previously discussed is suitable for modelling
the environmental footprints of conventional supervised, or one-time,
ML problems. On the other hand, novel learning paradigms are emerging
in industrial IoT setups. In the following sections, we highlight
some necessary adaptations with a particular focus on continual and
RL paradigms. Both scenarios are further discussed in the case studies
of Sect. \ref{subsec:Continual-and-few-shot}.

\textbf{Continual learning.} In continual learning paradigms the training
process must be periodically repeated to follow a time-varying industrial
process and using new input data or observations that change frequently.
When the input data changes, a new training process starts using either
the most recent global model instance as an initialization or meta-learning
tools \cite{maml} that quickly adapt to model changes or new tasks
\cite{IML}. Continual learning methods consist of an initial training
stage ($t_{0}$), where the model parameters are optimized over $n_{0}$
rounds using local training data $\mathcal{E}_{k}(t_{0})$, followed
by periodic tuning/re-training stages at subsequent time instants
$t_{1},...,t_{i}$. Model retraining uses a smaller amount of (few
shot) data \cite{simeone} as $b[\mathcal{E}_{k}(t_{i})]<b[\mathcal{E}_{k}(t_{0})]$,
and a reduced number of rounds, namely $n_{i}\ll n_{0},\forall i\geq1$.
Each stage has a cost of $E^{(t_{i})}(n_{i})$, typically with $E^{(t_{i})}(n_{i})\ll E^{(t_{0})}(n_{0})$
since fewer rounds are needed.

\textbf{Reinforcement learning (RL).} RL \cite{rl} is used to train
policies, typically deep ML models, that map observations of an environment
to a set of \emph{actions}, while trying to maximize a long-term \emph{reward}.
Similarly as for continual learning, the training process must be
periodically repeated, i.e., at time instants $t_{0},...,t_{i}$ using
new input observations $\mathcal{E}_{k}(t_{i})$, or \emph{states},
that change frequently as they are influenced by the same model under
training. In this paper, we resort to the popular Deep Q-Learning
(DQL) method \cite{deepmind}. The goal is to train a state-action
value function $\mathbf{W}$, namely the Q-function, that uses the
input states to predict the expected future rewards for each output
action, i.e., $\hat{y}\in\left\{ y_{c}\right\} _{c=1}^{C}$. The Q-function
is typically a deep ML model while training is obtained using gradient-based
optimization. Considering $K$ learning devices, the training process
on each round consists of new actions that generate new states/observations
of the environment: actions could be obtained by maximizing the local
Q-function $\mathbf{W}_{k,t}$ under training (exploitation) or they
can be randomly and independently chosen by devices (exploration).
The training data are therefore determined by the learned Q-function
and change at every round \cite{deepmind}. For CL, the training data
is moved to a data center on each round, therefore $\alpha=n$ in
(\ref{eq:cl}). On the other hand, FA, FA-D and CFA algorithms let
the devices exchange the parameters of their local Q-functions, rather
than the observations. All the algorithms proposed in Sects. \ref{subsec:Federated-Averaging}
and \ref{subsec:Consensus-driven-Federated-Avera} can be thus applied
without significant modifications. Finally, it is worth noticing that
since observations are collected on each new round, the energy footprint
must also include the cost of training data collection. A specific
example is given in Sect. \ref{subsec:Reinforcement-learning}.

\section{Design principles for low-carbon emissions \label{sec:Carbon-footprint-assessment}}

The section discusses the main factors that are expected to steer
the choice between centralized and federated learning paradigms towards
sustainable designs. Sustainability is here measured in terms of equivalent
GHG emissions, referred to as carbon footprints. The goal is to identify
the operating conditions that are necessary for FL policies (FA, FA-D
and CFA) to emit lower carbon than CL. Advantages of FL compared with
CL are related to communication and computing costs, as well as model
$b(\mathbf{W})$ and data $b(\mathcal{E}_{k})$ size. Considering
the energy models detailed in Sect. \ref{sec:Energy-footprint-modeling},
the carbon footprints (Sect. \ref{subsec:Energy-efficiency-and})
are obtained for all the FL policies ($\mathrm{C}_{\mathrm{FA}},\mathrm{C}_{\mathrm{FA-D}}$
and $\mathrm{C}_{\mathrm{CFA}}$) as well as for CL ($\mathrm{C}_{\mathrm{CL}}$).
Sustainable regions (Sect. \ref{subsec:Carbon-aware-design-constraints})
highlight necessary conditions on communication and computing energy
costs (energy and computing efficiencies) as well as on model v. data
footprint ratio $\frac{b(\mathbf{W})}{b(\mathcal{E}_{k})}$.

\subsection{Carbon footprints and model simplifications\label{subsec:Energy-efficiency-and}}

The carbon footprints $\mathrm{C}_{\mathrm{FA}},\mathrm{C}_{\mathrm{FA-D}},\mathrm{C}_{\mathrm{CFA}}$
and $\mathrm{C}_{\mathrm{CL}}$ are summarized in Tab. \ref{carbon_foots}
for all the proposed FL algorithms as well as for CL. Based on the
energy models (\ref{eq:cl})-(\ref{eq:cfa}), the estimated emissions
are evaluated by multiplying each individual energy contribution,
namely $E_{k}^{(\mathrm{C})}$ and $E_{k,h}^{(\mathrm{T})}$, by the
corresponding carbon intensity ($\mathrm{CI}_{k}$) of the electricity
generation \cite{LTE_power}. The $\mathrm{CI}_{k}$ terms depend
on the specific geographical regions where the devices $k=0,...,K$
are installed, and are measured in kg CO2-equivalent emissions per
kWh (kgCO2-eq/kWh): they quantify how much carbon emissions are produced
per kilowatt hour of locally generated electricity. 

Targeting general rules for sustainability assessment, the following
simplifications are applied to carbon footprints in Tab. \ref{carbon_foots}.
First, communication costs are quantified on average, in terms of
the corresponding energy efficiencies (EE), standardized by ETSI (European
Telecommunications Standards Institute \cite{ee_etsi}). These are
defined as the ratio between the data volume originated in DL ($\mathrm{EE}_{\mathrm{D}}=[E_{0,k}^{(\mathrm{T})}]^{-1}$),
UL ($\mathrm{EE}_{\mathrm{U}}=[E_{k,0}^{(\mathrm{T})}]^{-1}$) or
sidelink transmissions ($\mathrm{EE}_{\mathrm{S}}=[E_{k,h}^{(\mathrm{T})}]^{-1}$)
and the network energy consumption observed during the period required
to deliver the same data. Efficiency terms are measured here in bit/Joule
{[}bit/J{]} \cite{ee2,ee} and we consider different choices of $\mathrm{EE}_{\mathrm{D}}$,
$\mathrm{EE}_{\mathrm{U}}$ and $\mathrm{EE}_{\mathrm{S}}$ depending
on the specific network implementation. In particular, when the sidelink
interface is not available, but WWAN is used\footnote{It is worth mentioning that it would be $[\mathrm{EE}_{\mathrm{S}}]^{-1}=[\mathrm{EE}_{\mathrm{D}}]^{-1}+[\mathrm{EE}_{\mathrm{U}}]^{-1}$
if the PUE $\gamma$ of the Base Station or the router hardware were
$\gamma=1$ as mentioned in Sect. \ref{subsec:Federated-Averaging}. }, it is $[\mathrm{EE}_{\mathrm{S}}]^{-1}\simeq[\mathrm{EE}_{\mathrm{D}}]^{-1}+[\mathrm{EE}_{\mathrm{U}}]^{-1}$.
Considering now the computing costs, we define the computing efficiency
of the data center (or PS) as $\mathrm{EE}_{\mathrm{C}}=[E_{0}^{(\mathrm{C})}]^{-1}$.
It quantifies how much energy per learning round is consumed and it
is measured in terms of number of rounds per Joule {[}round/J{]}.
The computing efficiency of the devices $k>0$ are modeled here as
$[E_{k}^{(\mathrm{C})}]^{-1}=\frac{\mathrm{EE_{\mathrm{C}}}}{\varphi_{k}}$
with 
\begin{equation}
\varphi_{k}=\frac{E_{k}^{(\mathrm{C})}}{E_{0}^{(\mathrm{C})}}=\frac{P_{k}}{P_{0}}\cdot\frac{T_{k}}{T_{0}}.\label{eq:rho}
\end{equation}
Low-power devices typically experience a much larger local batch time
$T_{k}>T_{0}$ compared with data center $T_{0}$. On the other hand,
they use substantially lower power ($P_{k}\ll P_{0}$).

\begin{table*}[tp]
\begin{centering}
\protect\caption{\label{carbon_foots} Communication and computing carbon footprints.}
\vspace{-0.3cm}
\begin{tabular}{lll}
\toprule 
 & \textbf{Communication $\mathrm{C}^{(\mathrm{C})}$} \textbf{footprint} & \textbf{Computing $\mathrm{C}^{(\mathrm{L})}$} \textbf{footprint}\tabularnewline
\midrule 
\multirow{1}{*}{$\mathrm{C}_{\mathrm{CL}}$:}  & $\mathrm{C}_{\mathrm{CL}}^{(\mathrm{C})}=\alpha\cdot\sum_{k=1}^{K}b(\mathcal{E}_{k})\cdot\frac{\mathrm{CI}_{k}}{\mathrm{EE_{U}}}$  & $\mathrm{C}_{\mathrm{CL}}^{(\mathrm{L})}=n_{\mathrm{CL}}\cdot\gamma\cdot\frac{\mathrm{CI}_{0}}{\mathrm{EE_{C}}}$ \tabularnewline
\midrule 
$\mathrm{C}_{\mathrm{FA}}$: & $\mathrm{C}_{\mathrm{FA}}^{(\mathrm{C})}=n_{\mathrm{FA}}b(\mathbf{W})\left(\sum_{k=1}^{K_{a}}\frac{\mathrm{CI}_{k}}{\mathrm{EE_{U}}}+\gamma\cdot K\cdot\frac{\mathrm{CI}_{0}}{\mathrm{EE_{D}}}\right)$  & $\mathrm{C}_{\mathrm{FA}}^{(\mathrm{L})}=n_{\mathrm{FA}}\left(\sum_{k=1}^{K}\frac{\varphi_{k}\cdot\mathrm{CI}_{k}}{\mathrm{EE_{C}}}+\beta\cdot\gamma\cdot\frac{\mathrm{CI}_{0}}{\mathrm{EE_{C}}}\right)$ \tabularnewline
\midrule 
$\mathrm{C}_{\mathrm{FA-D}}$:  & $\mathrm{C}_{\mathrm{FA-D}}^{(\mathrm{C})}=n_{\mathrm{FA-D}}b(\mathbf{W})\left(\sum_{k=1}^{K_{a}}\frac{\mathrm{CI}_{k}}{\mathrm{EE_{U}}}+\gamma\cdot K_{a}\cdot\frac{\mathrm{CI}_{0}}{\mathrm{EE_{D}}}\right)$  & $\mathrm{C}_{\mathrm{FA-D}}^{(\mathrm{L})}=n_{\mathrm{FA-D}}\left(\sum_{k=1}^{K_{a}}\frac{\varphi_{k}\cdot\mathrm{CI}_{k}}{\mathrm{EE_{C}}}+\beta\cdot\gamma\cdot\frac{\mathrm{CI}_{0}}{\mathrm{EE_{C}}}\right)$ \tabularnewline
\midrule 
$\mathrm{C}_{\mathrm{CFA}}$:  & $\mathrm{C}_{\mathrm{CFA}}^{(\mathrm{C})}=n_{\mathrm{CFA}}b(\mathbf{W})\left(\sum_{k=1}^{K_{a}}\frac{N\cdot\mathrm{CI}_{k}}{\mathrm{EE_{S}}}\right)$  & $\mathrm{C}_{\mathrm{CFA}}^{(\mathrm{L})}=n_{\mathrm{CFA}}\sum_{k=1}^{K_{a}}\frac{\varphi_{k}\cdot\mathrm{CI}_{k}}{\mathrm{EE_{C}}}$ \tabularnewline
\bottomrule
\end{tabular}
\par\end{centering}
\medskip{}
 \vspace{-0.6cm}
\end{table*}

\subsection{Carbon-aware sustainable regions and requirements\label{subsec:Carbon-aware-design-constraints}}

Sustainability of FL depends on the specific operating conditions
about communication ($\mathrm{EE}_{\mathrm{D}}$, $\mathrm{EE}_{\mathrm{U}}$,
$\mathrm{EE}_{\mathrm{S}}$) and computing ($\mathrm{EE}_{\mathrm{S}}$)
efficiency, as well as model $b(\mathbf{W})$ and data $b(\mathcal{E}_{k})$
footprints. In what follows, we dive into such operational points
to highlight practical or necessary requirements for green designs.
To simplify the reasoning, we consider here $\mathrm{CI}_{k}\approx\mathrm{CI},\forall k$,
while the general results with arbitrary carbon intensity values $\mathrm{CI}_{k}$
are shown in the Appendix. All the requirements analyzed below are
summarized in Tab \ref{Region_definitions}.

\begin{table*}[tp]
\begin{centering}
\protect\caption{\label{Region_definitions} Region definitions and requirements.}
\vspace{-0.3cm}
\begin{tabular}{lll}
\toprule 
 & \textbf{Region definition} & \textbf{Requirement}\tabularnewline
\midrule 
\multirow{2}{*}{\multirow{1}{*}{$\mathcal{R}_{\mathrm{DU}}$:}} & $\left\{ \text{\ensuremath{\mathrm{EE}_{\mathrm{D}}},\ensuremath{\mathrm{EE}_{\mathrm{U}}}}:\mathrm{C}_{\mathrm{FA}}^{(\mathrm{C})}<\mathrm{C}_{\mathrm{CL}}^{(\mathrm{C})}\right\} $ & $\mathrm{\frac{EE_{D}}{\mathrm{EE}_{\mathrm{U}}}}\left(\frac{\alpha}{n\cdot K_{a}}\cdot\frac{\sum_{k=1}^{K}b(\mathcal{E}_{k})}{b(\mathbf{W})}-1\right)>\gamma\cdot\frac{K}{K_{a}}$\tabularnewline
 & $\left\{ \text{\ensuremath{\mathrm{EE}_{\mathrm{D}}},\ensuremath{\mathrm{EE}_{\mathrm{U}}}}:\mathrm{C}_{\mathrm{FA-D}}^{(\mathrm{C})}<\mathrm{C}_{\mathrm{CL}}^{(\mathrm{C})}\right\} $ & $\mathrm{\frac{EE_{D}}{\mathrm{EE}_{\mathrm{U}}}}\left(\frac{\alpha}{n\cdot K_{a}}\cdot\frac{\sum_{k=1}^{K}b(\mathcal{E}_{k})}{b(\mathbf{W})}-1\right)>\gamma$\tabularnewline
\midrule 
$\mathcal{R}_{\mathrm{SU}}$: & $\left\{ \text{\ensuremath{\mathrm{EE}_{\mathrm{S}}},\ensuremath{\mathrm{EE}_{\mathrm{U}}}}:\mathrm{C}_{\mathrm{CFA}}^{(\mathrm{C})}<\mathrm{C}_{\mathrm{CL}}^{(\mathrm{C})}\right\} $ & $\frac{\mathrm{EE_{S}}}{\mathrm{EE}_{\mathrm{U}}}\cdot\frac{\alpha}{n\cdot K_{a}}>N\cdot\frac{b(\mathbf{W})}{\sum_{k=1}^{K}b(\mathcal{E}_{k})}$\tabularnewline
\midrule 
\multirow{2}{*}{$\mathcal{R}_{b(\mathbf{W})}$:} & $\left\{ b(\mathbf{W}),b(\mathcal{E}_{k}):\max[\mathrm{C}_{\mathrm{FA}}^{(\mathrm{C})},\mathrm{C}_{\mathrm{FA-D}}^{(\mathrm{C})}]<\mathrm{C}_{\mathrm{CL}}^{(\mathrm{C})}\right\} $ & $\frac{b(\mathbf{W})}{b(\mathcal{E}_{k})}<\frac{\alpha}{n}\cdot\frac{K}{K_{a}}$\tabularnewline
 & $\left\{ b(\mathbf{W}),b(\mathcal{E}_{k}):\max[\mathrm{C}_{\mathrm{CFA}}^{(\mathrm{C})},\mathrm{C}_{\mathrm{FA}}^{(\mathrm{C})},\mathrm{C}_{\mathrm{FA-D}}^{(\mathrm{C})}]<\mathrm{C}_{\mathrm{CL}}^{(\mathrm{C})}\right\} $ & $\frac{b(\mathbf{W})}{b(\mathcal{E}_{k})}<\frac{\alpha}{n}\cdot\frac{K}{N\cdot K_{a}}$\tabularnewline
\midrule 
\multirow{3}{*}{$\mathcal{R}_{\mathrm{CI}}$:} & $\left\{ \text{\ensuremath{\mathrm{CI}_{k}}},\forall k:\frac{\mathrm{C}_{\mathrm{FA}}^{(\mathrm{L})}}{n_{\mathrm{FA}}}<\frac{\mathrm{C}_{\mathrm{CL}}^{(\mathrm{L})}}{n_{\mathrm{CL}}}\right\} $ & $\sum_{k=1}^{K}\frac{\varphi_{k}}{1-\beta}\mathrm{CI}_{k}<\gamma\mathrm{CI}_{0}$\tabularnewline
 & $\left\{ \text{\ensuremath{\mathrm{CI}_{k}}},\forall k:\frac{\mathrm{C}_{\mathrm{FA-D}}^{(\mathrm{L})}}{n_{\mathrm{FA-D}}}<\frac{\mathrm{C}_{\mathrm{CL}}^{(\mathrm{L})}}{n_{\mathrm{CL}}}\right\} $ & $\sum_{k=1}^{K_{a}}\frac{\varphi_{k}}{1-\beta}\mathrm{CI}_{k}<\gamma\mathrm{CI}_{0}$\tabularnewline
 & $\left\{ \text{\ensuremath{\mathrm{CI}_{k}}},\forall k:\frac{\mathrm{C}_{\mathrm{CFA}}^{(\mathrm{L})}}{n_{\mathrm{CFA}}}<\frac{\mathrm{C}_{\mathrm{CL}}^{(\mathrm{L})}}{n_{\mathrm{CL}}}\right\} $ & $\sum_{k=1}^{K}\varphi_{k}\mathrm{CI}_{k}<\gamma\mathrm{CI}_{0}$\tabularnewline
\bottomrule
\end{tabular}
\par\end{centering}
\medskip{}
 \vspace{-0.6cm}
\end{table*}
\textbf{Requirements on uplink and downlink efficiencies in cellular
communications.} FL policies make extensive use of UL/DL communications
either for local model parameters upload or global model download.
Quantifying necessary requirements on communication efficiencies $(\text{\ensuremath{\mathrm{EE}_{\mathrm{D}}},}\mathrm{EE}_{\mathrm{U}})$
is thus particularly critical and constitutes a key indicator for
the optimal choice between centralized and distributed learning. The
first problem we tackle is to identify the region of the parameter
space 
\begin{equation}
\mathcal{R}_{\mathrm{DU}}\triangleq\left\{ \text{\ensuremath{\mathrm{EE}_{\mathrm{D}}},\ensuremath{\mathrm{EE}_{\mathrm{U}}}}:\mathrm{C}_{\mathrm{FA}}^{(\mathrm{C})}<\mathrm{C}_{\mathrm{CL}}^{(\mathrm{C})}\right\} \label{eq:regiondl}
\end{equation}
such that when $\left\{ \mathrm{EE}_{\mathrm{D}},\mathrm{EE}_{\mathrm{U}}\right\} \in\mathcal{R}_{\mathrm{DU}}$
the FA policy emits lower carbon than CL with respect to communication
costs. Defining $b(\mathcal{E})=\sum_{k=1}^{K}b(\mathcal{E}_{k})$
as the size of the training data across the deployed devices, condition
$\mathrm{C}_{\mathrm{FA}}^{(\mathrm{C})}<\mathrm{C}_{\mathrm{CL}}^{(\mathrm{C})}$
in (\ref{eq:regiondl}) can be written as

\begin{equation}
\mathrm{\frac{EE_{D}}{\mathrm{EE}_{\mathrm{U}}}}\left(\frac{\alpha}{n\cdot K_{a}}\cdot\frac{b(\mathcal{E})}{b(\mathbf{W})}-1\right)>\gamma\cdot\frac{K}{K_{a}},\label{eq:downlink}
\end{equation}
with $\mathrm{C}_{\mathrm{FA}}^{(\mathrm{C})}$ and $\mathrm{C}_{\mathrm{CL}}^{(\mathrm{C})}$
in Tab. \ref{carbon_foots} (further details are in the Appendix).
Since training costs $\mathrm{C}^{(\mathrm{L})}$ are overlooked,
the bound (\ref{eq:downlink}) gives a necessary (but not sufficient)
condition towards sustainability assessment. As verified experimentally,
the bound (\ref{eq:downlink}) is revealed tight enough to serve as
a practical operating condition when communication emits much more
carbon than training, often verified in practice \cite{first_look,pimrc}.\textcolor{blue}{{}
}Considering FA-D, the parameter region (\ref{eq:regiondl}) becomes
$\mathcal{R}_{\mathrm{DU}}\coloneqq\left\{ \text{\ensuremath{\mathrm{EE}_{\mathrm{D}}},}\mathrm{EE}_{\mathrm{U}}:\mathrm{C}_{\mathrm{FA-D}}^{(\mathrm{C})}<\mathrm{C}_{\mathrm{CL}}^{(\mathrm{C})}\right\} $
while $\mathrm{C}_{\mathrm{FA-D}}^{(\mathrm{C})}<\mathrm{C}_{\mathrm{CL}}^{(\mathrm{C})}$
can be written as $\mathrm{\frac{EE_{D}}{\mathrm{EE}_{\mathrm{U}}}}\left(\frac{\alpha}{n\cdot K_{a}}\cdot\frac{b(\mathcal{E})}{b(\mathbf{W})}-1\right)>\gamma$.
Compared with (\ref{eq:downlink}), it gives a less stringent requirement
when $K_{a}<K$, namely the active device population $K_{a}$ is smaller
than the full population $K$. 

\textbf{Requirements on direct mode communications (sidelink efficiency).}
CFA does not need the PS and exploits direct mode communications,
replacing UL/DL communications with sidelinks (SL). Considering again
Tab. \ref{carbon_foots}, in analogy to (\ref{eq:regiondl}), we now
quantify necessary requirements on SL efficiency $\mathrm{EE_{S}}$.
The parameter region 
\begin{equation}
\mathcal{R}_{\mathrm{SU}}\triangleq\left\{ \text{\ensuremath{\mathrm{EE}_{\mathrm{S}}},\ensuremath{\mathrm{EE}_{\mathrm{U}}}}:\mathrm{C}_{\mathrm{CFA}}^{(\mathrm{C})}<\mathrm{C}_{\mathrm{CL}}^{(\mathrm{C})}\right\} \label{eq:regionsl}
\end{equation}
collects the operational points $\left\{ \mathrm{EE}_{\mathrm{S}},\mathrm{EE}_{\mathrm{U}}\right\} \in\mathcal{R}_{\mathrm{SU}}$
that make the CFA policy more efficient than CL in terms of communication
costs. Condition $\mathrm{C}_{\mathrm{CFA}}^{(\mathrm{C})}<\mathrm{C}_{\mathrm{CL}}^{(\mathrm{C})}$
in (\ref{eq:regionsl}) can be written as 
\begin{equation}
\frac{\mathrm{EE_{S}}}{\mathrm{EE}_{\mathrm{U}}}\cdot\frac{\alpha}{n\cdot K_{a}}>N\cdot\frac{b(\mathbf{W})}{b(\mathcal{E})}.\label{eq:EES}
\end{equation}
Similarly as (\ref{eq:downlink}), the bound (\ref{eq:EES}) gives
a necessary condition on SL efficiency. On the other hand, as verified
in Sect. \ref{subsec:Examples-with-MNIST}, it can be effectively
used for practical assessment when SL communication is the major source
of carbon emissions. As discussed in the Appendix, the condition $\frac{\mathrm{EE_{S}}}{\mathrm{EE_{U}}}+\gamma\frac{K}{K_{a}}\cdot\frac{\mathrm{EE_{S}}}{\mathrm{EE_{D}}}>N$
guarantees that, for each FL round, the CFA policy leaves a smaller
carbon footprint than FA, or FA-D (where $K$ is replaced with $K_{a}$).

\textbf{\textcolor{black}{Requirements on data and model size.}}\textcolor{blue}{{}
}In analogy to regions $\mathcal{R}_{\mathrm{DU}}$ and $\mathcal{R}_{\mathrm{SU}}$
that set the operational points for UL/DL and SL efficiency, the parameter
region
\begin{equation}
\mathcal{R}_{b(\mathbf{W})}\triangleq\left\{ b(\mathbf{W}),b(\mathcal{E}_{k}):\max[\mathrm{C}_{\mathrm{CFA}}^{(\mathrm{C})},\mathrm{C}_{\mathrm{FA}}^{(\mathrm{C})},\mathrm{C}_{\mathrm{FA-D}}^{(\mathrm{C})}]<\mathrm{C}_{\mathrm{CL}}^{(\mathrm{C})}\right\} .\label{eq:regionbw}
\end{equation}
identify the necessary requirements on model $b(\mathbf{W})$ and
data $b(\mathcal{E}_{k})$ size, now considering all the proposed
FL policies. Using (\ref{eq:downlink}) and (\ref{eq:EES}), the condition
$\max[\mathrm{C}_{\mathrm{CFA}}^{(\mathrm{C})},\mathrm{C}_{\mathrm{FA}}^{(\mathrm{C})},\mathrm{C}_{\mathrm{FA-D}}^{(\mathrm{C})}]<\mathrm{C}_{\mathrm{CL}}^{(\mathrm{C})}$
in $\mathcal{R}_{b(\mathbf{W})}$ can be written and simplified as
\begin{equation}
\frac{b(\mathbf{W})}{b(\mathcal{E}_{k})}<\frac{\alpha}{n}\cdot\frac{K}{\Gamma\cdot K_{a}}.\label{eq:size}
\end{equation}
by assuming $\frac{\mathrm{EE_{S}}}{\mathrm{EE}_{\mathrm{U}}},\mathrm{\frac{EE_{D}}{\mathrm{EE}_{\mathrm{U}}}}\gg1$
\cite{LTE_power} and $\forall k,\,b(\mathcal{E}_{k})=b(\mathcal{E})/K$
as clarified in the Appendix. In particular, taking into consideration
FA and FA-D policies only, it is $\Gamma=1$, while $\Gamma=N$ when
including CFA. Notice that, for an assigned model $b(\mathbf{W})$
and dataset $b(\mathcal{E}_{k})$ size, a critical requirement for
(\ref{eq:size}) to hold is minimizing the active population size
($K_{a}<K$) and keeping the number of learning rounds ($n$) to a
minimum, while satisfying a target accuracy. Furthermore, frequent
data updates, as often observed in continual learning ($\alpha>1$),
might favor FL. 

\textbf{\textcolor{black}{Requirements on computing efficiency.}}\textcolor{blue}{{}
}Considering now the training costs, the parameter region 
\begin{equation}
\mathcal{R}_{\mathrm{CI}}\triangleq\left\{ \text{\ensuremath{\mathrm{CI}_{k}}},\forall k:\frac{\mathrm{C}_{\mathrm{FA}}^{(\mathrm{L})}}{n_{\mathrm{FA}}}<\frac{\mathrm{C}_{\mathrm{CL}}^{(\mathrm{L})}}{n_{\mathrm{CL}}}\right\} \label{eq:regionci}
\end{equation}
sets the requirements on carbon intensities such that, for $\text{\ensuremath{\mathrm{CI}_{k}}}\in\mathcal{R}_{\mathrm{CI}}$,
the FA policy emits lower carbon than CL with respect to the per-round
computing costs. After straightforward manipulation of $\mathrm{C}_{\mathrm{FA}}^{(\mathrm{L})}$
and $\mathrm{C}_{\mathrm{CL}}^{(\mathrm{L})}$ from Tab. \ref{carbon_foots},
the bound $\frac{\mathrm{C}_{\mathrm{FA}}^{(\mathrm{L})}}{n_{\mathrm{FA}}}<\frac{\mathrm{C}_{\mathrm{CL}}^{(\mathrm{L})}}{n_{\mathrm{CL}}}$
in (\ref{eq:regionci}) can be written as 
\begin{equation}
\sum_{k=1}^{K}\frac{\varphi_{k}}{1-\beta}\mathrm{CI}_{k}<\gamma\mathrm{CI}_{0}.\label{eq:training}
\end{equation}
Extending requirement (\ref{eq:training}) to other FL policies, namely
FA-D, $\frac{\mathrm{C}_{\mathrm{FA-D}}^{(\mathrm{L})}}{n_{\mathrm{CFA}}}<\frac{\mathrm{C}_{\mathrm{CL}}^{(\mathrm{L})}}{n_{\mathrm{CL}}}$,
and CFA, $\frac{\mathrm{C}_{\mathrm{CFA}}^{(\mathrm{L})}}{n_{\mathrm{CFA}}}<\frac{\mathrm{C}_{\mathrm{CL}}^{(\mathrm{L})}}{n_{\mathrm{CL}}}$,
is straightforward. The corresponding requirements are detailed in
the Table \ref{Region_definitions}.

\section{Carbon footprints: examples in cellular networks\label{subsec:Examples-with-MNIST}}

In this section we verify the sustainability conditions (\ref{eq:downlink})-(\ref{eq:training})
and the regions in Table \ref{Region_definitions} for typical supervised
image classification problems ($\alpha=1$), using the popular MNIST
\cite{mnist} and CIFAR10 \cite{cifar10} datasets. In line with 4G/5G
NB-IoT cellular network implementations \cite{survey_energy_5G},
model parameters and data exchange are here implemented over a WWAN
characterized by UL $\mathrm{EE_{U}}=10$ kbit/J and DL efficiency
$\mathrm{EE_{D}}=50$ kbit/J. Other setups are considered in Sect.
\ref{sec:A-case-study}. Three populations of devices are analyzed,
ranging from $K=100$, of which $K_{a}=50$ are active learners, $K=60$
(with $K_{a}=40$) and $K=30$ ($K_{a}=20$). For these initial tests,
distributed learning has been simulated on a framework that allows
to deploy virtual devices implemented as independent threads running
on the same machine. Each thread acts as learner and process a fraction
of an assigned data set while communicating with the other threads
through a resource sharing system \cite{cfa}. The simulations allow
to compute the number of rounds $n$ that are necessary to achieve
a target accuracy or loss $\overline{\xi}$: energy and carbon footprints
are then obtained for each setup by evaluating communication and computing
costs in Tab. \ref{carbon_foots}. For CL consumption, the data center
machine is equipped with a RTX 3090 GPU hardware (with Thermal Design
Power equal to $350$W) having PUE $\gamma=1.67$. We consider the
carbon intensity $\mathrm{CI}_{k}$ figures reported in EU back in
2019 \cite{CI}: in particular, $\mathrm{CI}_{k}=0.9$ kgCO2-eq/kWh,
according to \cite{trend} (mlco2.github.io/impact/). 

The following analysis follows a what-if approach to quantify the
estimated carbon emissions under different parameter choices. Actual
emissions may be larger than the estimated ones depending on the specific
use case and network implementation: relative comparisons are however
meaningful for green design assessment. Code examples are available
online\footnote{Federated Learning code repository: https://github.com/labRadioVision/federated.
Accessed: March 2022.}.

\begin{table*}[tp]
\protect\caption{\label{rounds_all} FL rounds for varying target losses $\overline{\xi}$,
and IID vs non-IID data distributions with $\mathrm{EE}_{\mathrm{U}}=10$
kbit/J, $\mathrm{EE}_{\mathrm{D}}=50$ kbit/J. Examples with the MNIST
and the CIFAR10 datasets are shown as well.}
\vspace{-0.3cm}
 \centering \includegraphics[scale=0.34]{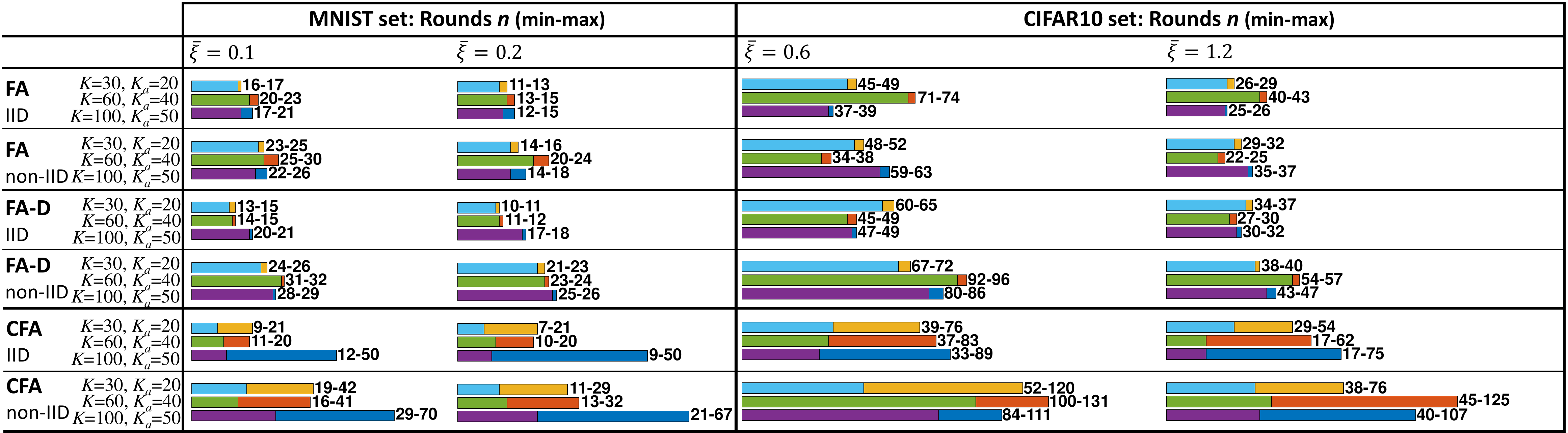}
 \medskip{}
 \vspace{-0.6cm}
\end{table*}
\begin{table*}[tp]
\protect\caption{\label{mnist} Communication/computing energy costs and corresponding
carbon footprints increase/decrease w.r.t. CL for the MNIST dataset.}
\vspace{-0.3cm}
 \centering \includegraphics[scale=0.36]{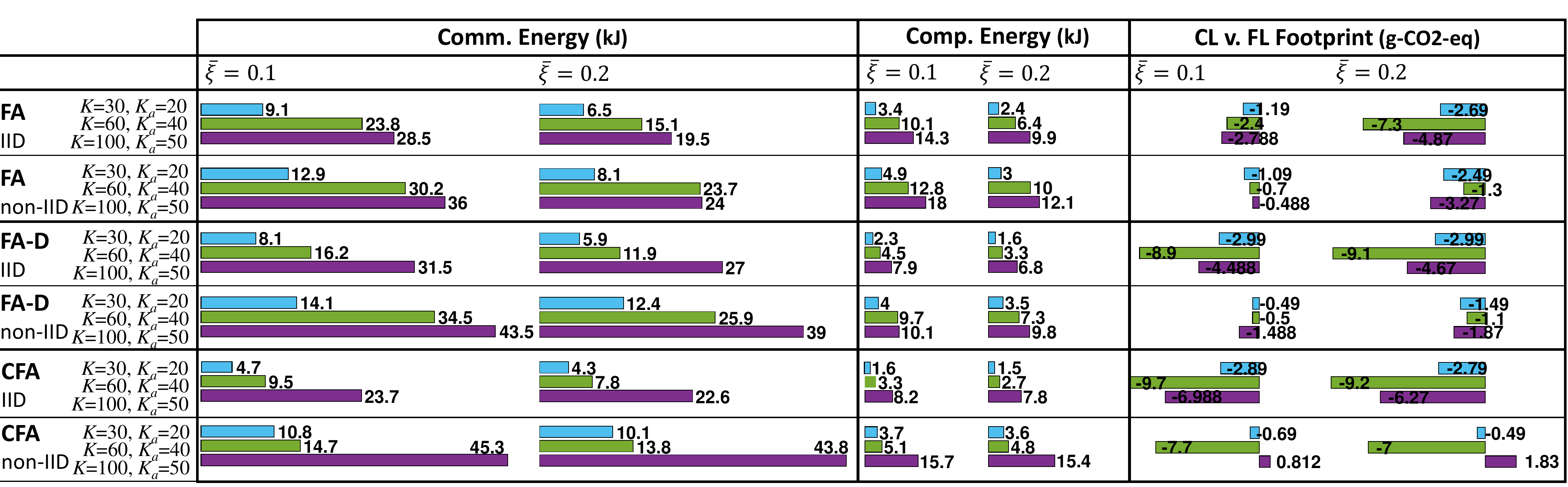} 
 \medskip{}
 \vspace{-0.6cm}
\end{table*}

\subsection{Impact of conditions (\ref{eq:downlink})-(\ref{eq:training}) on
MNIST and CIFAR\label{subsec:Impact-of-conditions}}

Handwritten image recognition from the MNIST dataset uses the LeNet-1
model proposed in \cite{mnist}. Each device obtains $1000$ MNIST
training gray scale images with average (per device) data footprint
$b(\mathcal{E}_{k})=6.2$ Mbit, $\forall k$. Model size with no compression
is $b(\mathbf{W})=180$ kbit, so that $\frac{b(\mathbf{W})}{b(\mathcal{E}_{k})}\simeq0.03$.
Considering the CIFAR set, devices obtain $1000$ CIFAR10 color images,
$b(\mathcal{E}_{k})=24.5$ Mbit, and use a VGG-1 model with $b(\mathbf{W})=2$
Mbit, so that $\frac{b(\mathbf{W})}{b(\mathcal{E}_{k})}\simeq0.08$.
In the following tests, both IID and non-IID data distributions are
considered: in particular for non-IID data, batches now contain examples
for $6$ of the $10$ target classes (MNIST or CIFAR), randomly chosen. 

\textbf{MNIST carbon footprints}. Tab. \ref{rounds_all} (left column)
quantifies the minimum and maximum number of rounds ($n$) required
for FA, FA-D and CFA, as targeting different validation loss values
$\overline{\xi}$. Communication, computing energy costs, and carbon
footprints are instead reported in Tab. \ref{mnist}. The carbon footprint
is quantified by decrease (negative terms) or increase (positive terms)
with respect to the emissions observed for CL, used here as reference.
The carbon emissions of CL are found as $17.6$ gCO2-eq for $K=100$
devices, $14.1$ gCO2-eq for $K=60$, and $5.6$ gCO2-eq for $K=30$.
The parameter region $\mathcal{R}_{b(\mathbf{W})}$ and the upper
bound (\ref{eq:size}) can be used to quantify the maximum number
of rounds below which distributed learning tools could be considered
as sustainable: for $\frac{b(\mathbf{W})}{b(\mathcal{E}_{k})}\simeq0.03$,
and after straightforward manipulation, it is $n<50$ for $K=30$
and $K=60$, and $n<66$ for $K=100$. FA and FA-D methods both satisfy
the previous condition as they require $n=13\sim29$ rounds for $\overline{\xi}=0.1$
and fewer rounds $n=11\sim26$ for $\overline{\xi}=0.2$. The region
$\mathcal{R}_{\mathrm{DU}}$ and condition (\ref{eq:downlink}) for
$n=29$ rounds, $K=30$, and $K_{a}=20$ indicates $\mathrm{\frac{EE_{D}}{\mathrm{EE}_{\mathrm{U}}}}>2.8$
for FA and $\mathrm{\frac{EE_{D}}{\mathrm{EE}_{\mathrm{U}}}}>1.9$
for FA-D that are both satisfied as $\mathrm{\frac{EE_{D}}{\mathrm{EE}_{\mathrm{U}}}}=5$.
CFA leaves a smaller carbon footprint than CL for IID data distributions.
However, when data is non-IID, it requires up to $n=70$ rounds with
$K=100$ devices that exceeds the bound (\ref{eq:size}) and thus
cause a larger footprint. Notice that, for CFA, the observed rounds
might differ substantially over varying simulation runs, as the result
of the propagation of model parameters through the network. 

\textbf{CIFAR carbon footprints}. Tab. \ref{rounds_all} (right column)
and Tab. \ref{cifar} collect the required rounds, the energy and
carbon footprints for the CIFAR set, respectively. Classification
with CIFAR requires much more rounds for convergence than MNIST and
a larger model footprint $b(\mathbf{W})$: according to the region
$\mathcal{R}_{b(\mathbf{W})}$ and the bound (\ref{eq:size}), FL
is less attractive with respect to centralized designs. For CL, the
number of rounds to achieve $\overline{\xi}=0.6$ (accuracy of $77$\%)
is in the range $n_{\mathrm{CL}}=30\sim40$, while the carbon footprint
is found to be $267.4$ gCO2-eq for $K=100$ devices, $161.5$ gCO2-eq
for $K=60$, and $81.4$ gCO2-eq for $K=30$, respectively. Tab. \ref{cifar}
shows that all FL paradigms leave larger footprints than CL since
the required number of rounds exceeds the bound (\ref{eq:size}),
that prescribes a maximum of $25$ rounds. A sustainable solution
could be achieved only in exchange for a larger loss: in other words,
increasing the target loss to $\overline{\xi}=1.2$ reduces the required
number of FL rounds, but penalizes the accuracy that scales down to
$67\%$. This last setup might give concrete chances to CFA, when
data is IID distributed, considering the lower communication cost
per round as shown in the requirement (\ref{eq:EES}) or region $\mathcal{R}_{\mathrm{SU}}$.
Finally, for $\mathrm{CI}_{k}=0.9$, $\varphi_{k}=0.14$ and $\beta=0.05$,
condition (\ref{eq:training}) on training costs is not satisfied:
therefore, FL emits much more carbon than CL on each learning round
as $\text{\ensuremath{\mathrm{CI}_{k}}}\notin\mathcal{R}_{\mathrm{CI}}$. 

\begin{table*}[tp]
\protect\caption{\label{cifar} Communication/computing energy costs and corresponding
carbon footprints increase/decrease w.r.t. CL for the CIFAR10 dataset.}
\vspace{-0.3cm}
 \centering \includegraphics[scale=0.36]{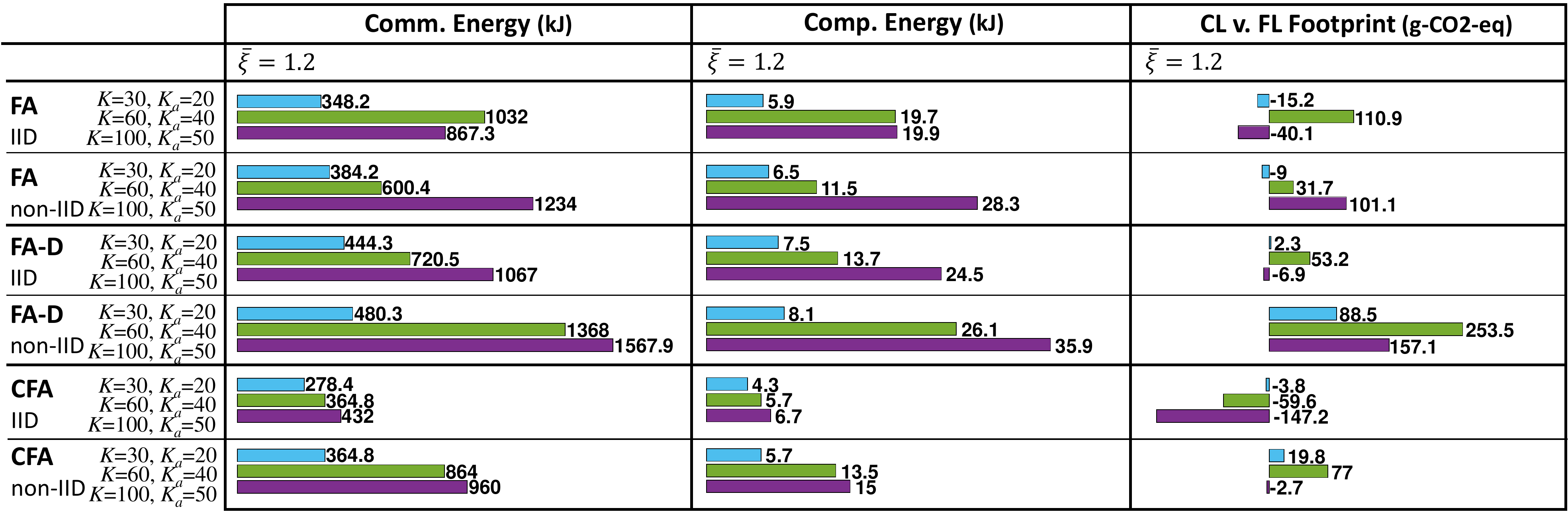} 
 \medskip{}
 \vspace{-0.6cm}
\end{table*}

\subsection{MNIST and CIFAR: comparative analysis\label{subsec:MNIST-and-CIFAR:}}

\begin{figure}[!t]
\centering \includegraphics[scale=0.65]{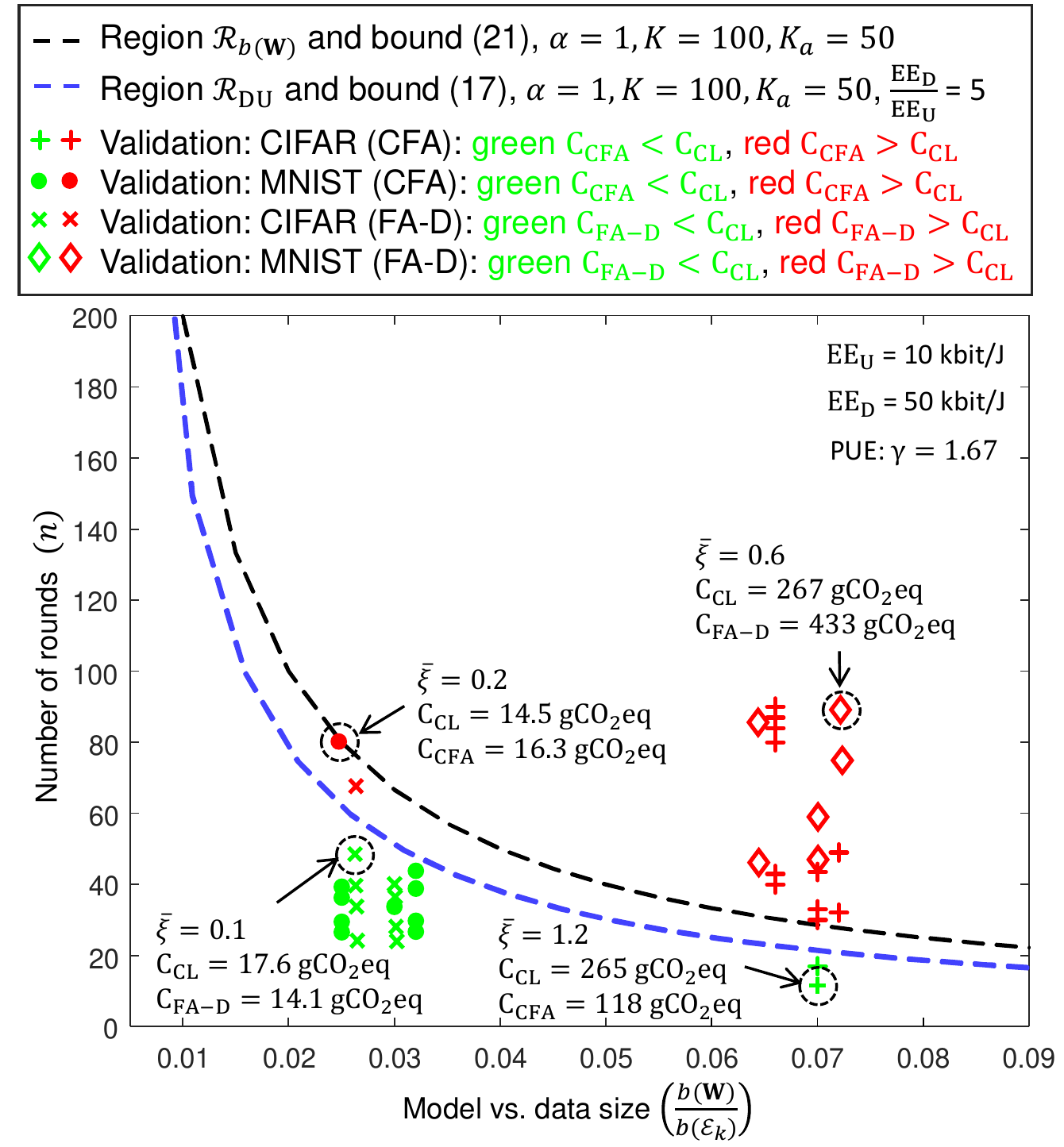} 
 \protect\caption{\label{bound_validation} Validation of the bounds (\ref{eq:size})
in black, and of (\ref{eq:downlink}) in blue, with the CIFAR and
the MNIST datasets for the CFA and FA-D methods. Markers in the scatter
plot report FA-D or CFA simulation runs for the MNIST or CIFAR datasets
(IID or non-IID), and give the required rounds for the target loss
$\overline{\xi}$ and the corresponding model/data size $\frac{b(\mathbf{W})}{b(\mathcal{E}_{k})}$.
Green markers are sustainable solutions, namely $\mathrm{C}_{\mathrm{FA-D}}<\mathrm{C}_{\mathrm{CL}}$
or $\mathrm{C}_{\mathrm{CFA}}<\mathrm{C}_{\mathrm{CL}}$, while red
markers refers to the opposite cases. }
\vspace{-0.5cm}
\end{figure}

By comparing the MNIST and the CIFAR sets, some common properties
emerge: these are conveniently summarized in Fig. \ref{bound_validation},
focusing on FA-D and CFA methods. Each point in the scatter plot corresponds
to a simulation run that uses FA-D or CFA over the MNIST or the CIFAR
sets. For each combination, represented by different markers, we run
several simulations by varying the data distribution, from IID to
non-IID\footnote{Non-IID data footprint $b(\mathcal{E}_{k})$ might vary in each simulation
run, depending on the training sample distribution.}, and we report the required number of rounds, for accuracy $\overline{\xi}$,
as well as the corresponding model/data footprint ratio $\frac{b(\mathbf{W})}{b(\mathcal{E}_{k})}$.
The bounds (\ref{eq:size}) and (\ref{eq:downlink}) are superimposed
by black and blue dashed lines, respectively: the area above each
line represents unsustainable designs, namely $\left\{ b(\mathbf{W}),b(\mathcal{E}_{k})\right\} \notin\mathcal{R}_{b(\mathbf{W})}$
and $\left\{ \mathrm{EE}_{\mathrm{D}},\mathrm{EE}_{\mathrm{U}}\right\} \notin\mathcal{R}_{\mathrm{DU}}$,
for which CL is the preferred choice, with respect to carbon emissions.\textcolor{blue}{{}
}In most cases, the simulations confirm the predicted trends: green
markers correspond to sustainable solutions for which FA-D, or CFA,
leave a lower carbon footprint than CL, red markers refer to the opposite
case. Small model footprints $b(\mathbf{W})$ compared with data $b(\mathcal{E}_{k})$
constitute a critical prerequisite for FL sustainability, while learning
rounds ($n$) must be minimized as much as possible. Notice that the
required rounds increase when the data is non-IID. CFA is generally
more sensitive to non-IID distributions than FA-D; on the other hand,
in many cases, CFA gives better footprints in IID situations. Regions
$\mathcal{R}_{\mathrm{DU}}$, $\mathcal{R}_{b(\mathbf{W})}$ and corresponding
requirements (\ref{eq:downlink})-(\ref{eq:size}) give more effective
indicators of sustainability than region $\mathcal{R}_{\mathrm{CI}}$
and bound (\ref{eq:training}) on training costs. This is due to the
fact that, in these examples, communication emits much more carbon
than computing.

\begin{figure}[!t]
\centering \includegraphics[scale=0.32]{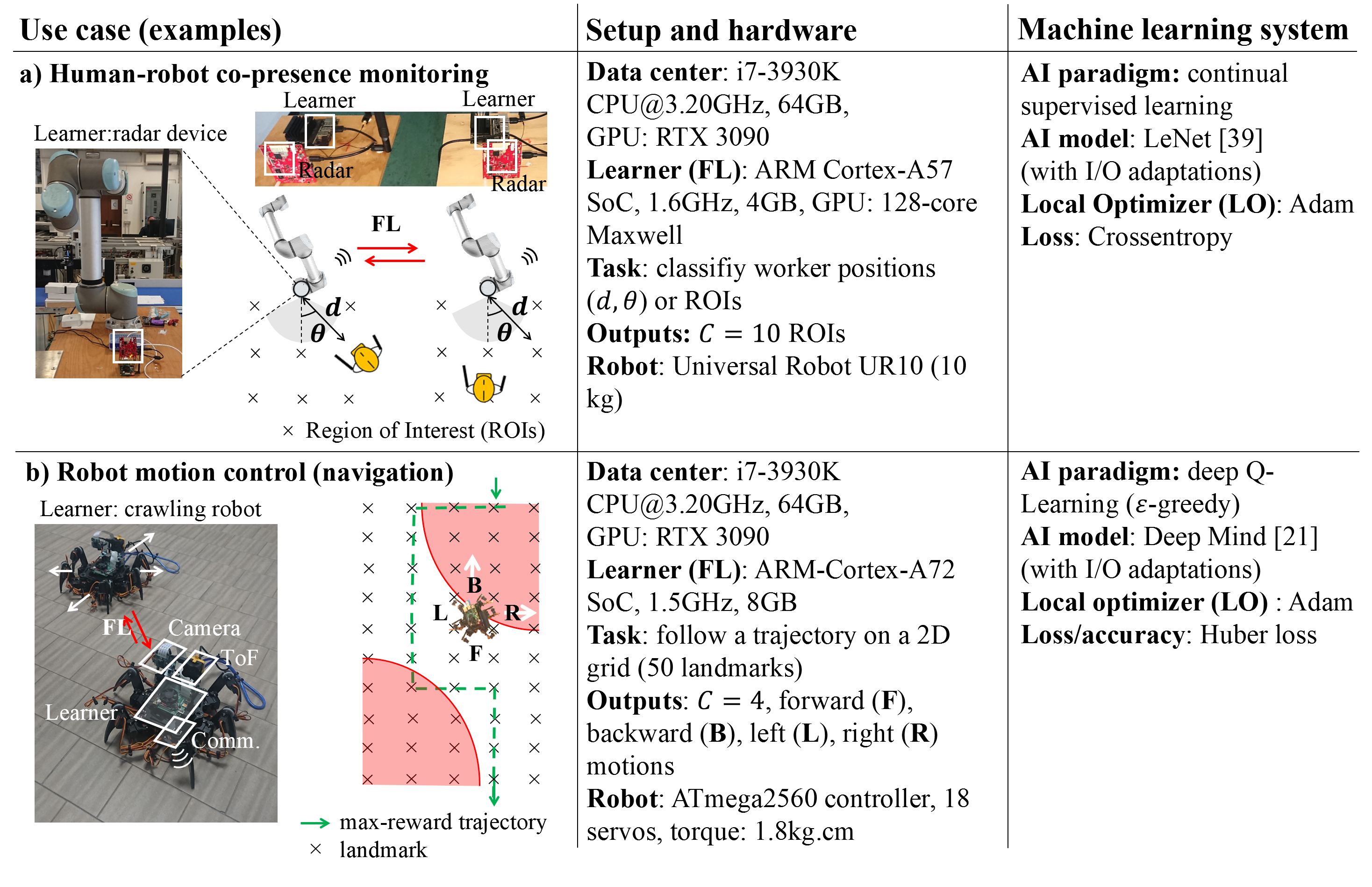} 
 \protect\caption{\label{table_usecases} Description of case studies. From left to
right: scenario descriptions, hardware specifications for each industrial
setup and corresponding ML systems.}
\vspace{-0.5cm}
\end{figure}

\section{Case studies in industrial IoT \label{sec:A-case-study}}

In what follows, we highlight two critical 5G verticals\footnote{These case studies have been taken from 5G Alliance for Connected
Industries and Automation: https://5g-acia.org/} for Industry 4.0 scenarios, further described in Fig. \ref{table_usecases}.
The interested reader might refer to \cite{link_sust_industry40}
for sustainability requirements in industry processes and transition
towards the new Green Deal \cite{green_deal}. 

The first use case, shown in Fig. \ref{table_usecases}(a), quantifies
the carbon footprint of a continual supervised learning process targeting
passive localization of human operators in a shared workplace (human-robot
co-presence monitoring). Localization is particularly critical to
support various human-robot interactions processes \cite{kianoush}
in advanced manufacturing, where robotic manipulators continuously
share the same space with humans. A deep ML model is adopted for localization
and it is trained continuously to track the variations of data dynamics
caused by changes of the workflow processes (typically, on a daily
base). In the second setup, described in Fig. \ref{table_usecases}(b),
$K=5$ crawling robots interact with the workplace to learn an optimal
sequence of movements and implement a desired trajectory. Robots train
a ML model, now via DQL tools \cite{deepmind} as introduced in Sect.
\ref{subsec:Adaptations-to-continual,}. Both case studies focus on
industrial setups where AI-based machines/robots are co-located in
the same workplace, so that direct communication is possible. 

For all case studies, it is assumed that devices have sufficient computing
power to implement FL and LO, but they run medium/small-sized deep
learning models to allow for real-time operations, i.e., localization
or motion control. Hardware and ML systems are described in Fig. \ref{table_usecases}.
The devices, the PS (when used), and the data centers are located
in the west EU regions previously defined. In particular, the data
center, and the PS are located outside the workplace so that communication
is possible only through WWAN connectivity. They are equipped with
the same high performance GPU already shown in Sect. \ref{subsec:Examples-with-MNIST}
and have the same PUE namely $\gamma=1.67$. However, a realistic
pool of FL learners is considered: manipulator devices are equipped
with dedicated hardware modules (Nvidia Jetson Nano) equipped with
a low-power GPU (128-core Maxwell architecture) and an ARM Cortex-A57
System-on-Chip (SoC). Crawling robots mount a low-power ARM Cortex-A72
SoC and thus experience larger batch time $T_{k}$ but lower power
$P_{k}$ when implementing DQL. 

As described in the following, the proposed real-time FL platform
adopts MQTT methods for model parameter exchange and several communication
profiles. The carbon footprints for both case studies are analyzed
separately, while sustainable setups are identified based on the requirements
in Table \ref{Region_definitions}. 
\begin{table}[tp]
\protect\caption{\label{parameters} Summary of main FL parameters for use cases.}
\vspace{-0.2cm}

\begin{centering}
\begin{tabular}{l|l|l||l|l|}
\cline{2-5} 
 & \multicolumn{2}{c||}{\textbf{Case (a): Continual Learning}} & \multicolumn{2}{c|}{\textbf{Case (b): Reinforcement Learning}}\tabularnewline
\hline 
$b\left[\mathcal{E}_{k}(t_{i})\right]$: & \multicolumn{2}{l||}{$31$ MB ($t_{0}$), $19$ MB ($t_{i}$), $\forall k,i$} & \multicolumn{2}{l|}{$24.6$ MB, $\forall k,i$}\tabularnewline
\hline 
$b(\mathbf{W})$: & \multicolumn{2}{l||}{$1.08$ MB} & \multicolumn{2}{l|}{$5.6$ MB}\tabularnewline
\hline 
$\alpha$: & \multicolumn{2}{l||}{$1$ (1 updates per training, $\forall t_{i}$)} & \multicolumn{2}{l|}{$\text{\ensuremath{n}}$ ($1$ update per round)}\tabularnewline
\hline 
$K,K_{a}:$ & \multicolumn{2}{l||}{$K=9,K_{a}:1\sim9$} & \multicolumn{2}{l|}{$K=5,K_{a}=5$}\tabularnewline
\hline 
\textbf{Consumption} & \textbf{Data center/PS ($k=0$)} & \textbf{Devices ($k\geq1$)} & \textbf{Data center/PS ($k=0$)} & \textbf{Devices ($k\geq1$)}\tabularnewline
\hline 
$P_{k}$: & $590\,\mathrm{W}$$\,\,(350\,\mathrm{W}$ GPU) & $15\,\mathrm{W}$$(\varphi_{k}=0.36)$ & $590\,\mathrm{W}$$\,\,(350\,\mathrm{W}$ GPU) & $5.1\,\mathrm{W}$ $(\varphi_{k}=0.17)$\tabularnewline
\hline 
$T_{k}$: & $10\textrm{ms}$ ($B=3$) & $140\textrm{ms}$ ($B=3$) & $20\textrm{ms}$ ($B=3$) & $400\textrm{ms}$ ($B=3$)\tabularnewline
\hline 
$\beta$: & $0.05$ & n.a. & $0.05$ & n.a.\tabularnewline
\hline 
$E_{k,\mathrm{sleep}}^{(\mathrm{C})}:$ & n.a. & $0.05$ Wh & n.a. & $0.05$ Wh\tabularnewline
\hline 
$\mathrm{EE_{C}:}$ & $0.05$ round/J & $0.14$ & $\mathrm{EE_{C}}=0.03$ round/J & $0.16$\tabularnewline
\hline 
\hline 
\textbf{Peripherals} & n.a. & $3$ Wh (radars) & n.a. & $6.6$ Wh (servos)\tabularnewline
\hline 
\end{tabular}
\par\end{centering}
\medskip{}
 \vspace{-0.6cm}
\end{table}

\subsection{Networking and MQTT transport\label{subsec:Networking-and-MQTT}}

The networking platform used for validation is characterized by $K$
physical IoT devices and implements distributed model training in
real-time on top of the MQTT protocol, chosen for low-latency and
low-overhead characteristics. For all FL implementations, the exchange
of model parameters is performed through MQTT-compliant \emph{publish}
and \emph{subscribe} operations that are implemented with QoS level
1 \cite{wot-sw}. The information included in the MQTT payload are:
\emph{i)} the local model parameters binary encoded with associated
meta-information (i.e., the DNN layer type); \emph{ii)} the FL round;
\emph{iii)} the local loss function used as model quality indicator.
The MQTT platform is used to collect measurements of the required
training time: the expected carbon footprints are then quantified
for varying communication energy efficiencies ($\mathrm{EE_{D}},$
$\mathrm{EE_{U}},$ $\mathrm{EE_{S}}$). The code structure is available
in \cite{dataport}. CL is similarly implemented on top of the same
MQTT architecture: in this case, the $K$ devices publish their data
to the MQTT broker that is co-located with the data center. 

Considering FA and FA-D methods, the PS serves as MQTT broker until
the end of the training process: on each FL round, the PS thus accepts
subscriptions from the $K_{a}$ active devices that publish their
model parameters. For FA method, all the $K$ devices subscribe to
the PS broker service, i.e., to download the updated global model.
On the other hand, for FA-D, inactive devices need to start a new
subscription process every time they wake up from deep-sleep. The
$K_{a}$ learners are chosen by a round robin scheduling table \cite{energy2}
that is distributed by the PS at training start, via a dedicated MQTT
topic. The broker service might cause an increase of the consumption
per round, or a decrease of the computing efficiency $\mathrm{EE_{C}}$.
Similarly, subscriptions and publishing operations reduce the communication
efficiency, $\mathrm{EE_{D}},$ $\mathrm{EE_{U}}$ depending on the
payload-to-overhead ratio of the MQTT messaging \cite{wot-sw}. To
implement CFA, the MQTT broker is now used to orchestrate the consensus
operations, while the PS functions are disabled. Every round, one
device is turned on and publishes the local model parameters to a
subset of $K_{a}-1$ subscriber devices. Publisher and subscribers
are again assigned by round robin scheduling. Notice that the MQTT
broker consumption is not considered in this study. 

\subsection{Communication efficiencies\label{subsec:Communication-efficiencies}}

The estimated carbon emissions are quantified by considering $4$
communication efficiency profiles \cite{survey_energy_5G} described
in the following. The first profile conforms to a LTE design for macro-cell
delivery: we set $\mathrm{EE_{U}}=15$ kbit/J and $\mathrm{EE_{D}}=25$
kbit/J as also verified in \cite{ee} for a throughput of $28$ Mbps
and a BS consumption of $1.35$ kW. Notice that larger efficiency
values of $6$ J/Mbit and $10$ J/Mbit for UL and DL, respectively,
could be observed with small bulk size of $10$ kB \cite{LTE_power}.
The second profile corresponds to a NB-IoT implementation. In this
case, we set a larger DL $\mathrm{EE_{D}}=50$ kbit/J and UL efficiency
of $\mathrm{EE_{U}}=25$ kbit/J\footnote{In \cite{nbiot}, it is shown that sending $20$ byte using $0.9$
Joule is possible: efficiency reduces with larger packet sizes. }. NB-IoT devices also consume about $10$ \textmu W (i.e., $E_{k,\mathrm{sleep}}^{(\mathrm{C})}\simeq36$
mWh) in deep-sleep mode \cite{nbiot}. The third profile complies
with IETF 6TiSCH mesh standard based on IEEE 802.15.4e \cite{tisch}
with a SL efficiency of $\mathrm{EE_{S}}=20$ kbit/J. For the last
profile, we adopt a WiFi IEEE 802.11ac implementation, namely the
Intel AC 8265 device supporting sidelink connectivity based on Neighbor
Awareness Network \cite{wifiaware}. Typical communication efficiencies
are $2100$ mW/Mbps for receiving and $2500$ mW/Mbps for transmitting
\cite{d2dcomm}; however, they do not include equipment consumption,
nor WiFi and MQTT overheads: the assumed SL efficiency is thus $\mathrm{EE_{S}}=100$
kbit/J. Notice that sidelink efficiency is expected to scale up to
at least $10$ times in 6G implementations \cite{survey_energy_5G}.

\subsection{Human-robot interaction: continual learning \label{subsec:Continual-and-few-shot}}

The goal of the training task is to learn a ML model for the detection
(i.e., classification) of the position of the human operators sharing
the workspace, namely the human-robot distance $d$ and the direction
of arrival (DOA) $\theta$. In particular, we address the detection
of a human subject in $C=10$ Region Of Interest (ROI), including
the one referring to the subject outside the monitored area: ROIs
are detailed in Fig. \ref{table_usecases}(a). The proposed training
scenario resorts to a network of $K=9$ physical devices where each
one is equipped with a Time-Division Multiple-Input-Multiple Output
(TD-MIMO) Frequency Modulated Continuous Wave (FMCW) radar working
in the $77-81$ GHz band. For details about the robotic manipulators,
the industrial environment and the radars, the interested reader may
also refer to \cite{kianoush}. Radars use a trained deep learning
model to obtain position ($d$, $\theta$) information and the corresponding
ROI. In addition, the subject position can be sent to a programmable
logic controller for robot safety control, for emergency stop or replanning
tasks. The ML model adopted for the classification of the operator
location is the LeNet-4 scheme, proposed in \cite{mnist} with input
adaptations, that consists of $4$ trainable layers and $28$K parameters.
Model footprint is $b(\mathbf{W})=1.08$ MB, the other parameters
are detailed in Tab. \ref{parameters}. 

The FL system implements a continual learning task (Sect. \ref{subsec:Adaptations-to-continual,}):
at time $t_{0}$, each device collects a large dataset $\mathbf{x}_{h}$
of raw range-azimuth data manually labelled, with size $b[\mathcal{E}_{k}(t_{0})]=31$
MB. This set is used for initial training of the ML model. Re-training
stages, $t_{i}$, $i>0$, are based on new data, $b[\mathcal{E}_{k}(t_{i})]=19$
MB, collected on a daily basis. Datasets for initial model training
and $2$ subsequent example re-training stages, $t_{1},t_{2}$, are
available online in \cite{dataport}. To simplify the analysis, in
what follows, we compared the results of the proposed FA-D and CFA
implementations, while FA is not considered. For FA-D, the number
of active devices ranges from $K_{a}=1$ to $K_{a}=9$; for CFA, we
have assumed $K_{a}=3$ ($1$ publisher, $2$ subscribers per round)
up to $K_{a}=9$ ($1$ publisher, $8$ subscribers per round).
\begin{figure}[!t]
\centering \includegraphics[scale=0.44]{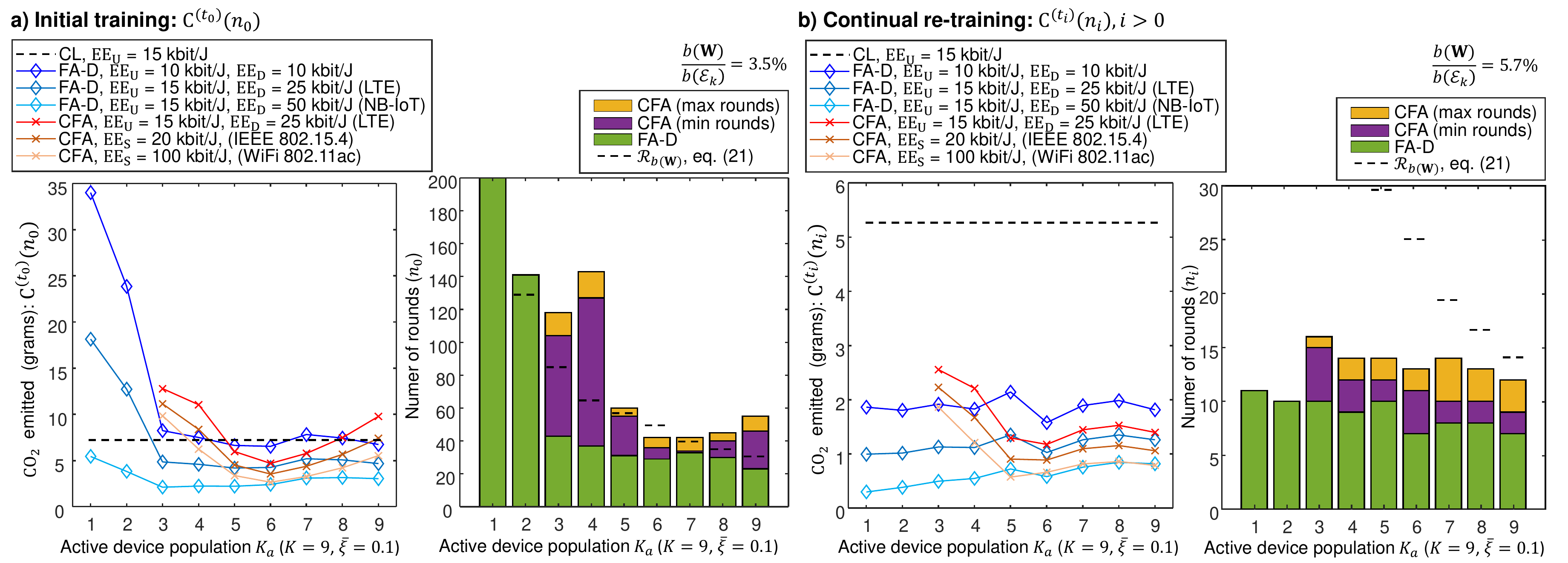}
 \protect\caption{\label{comp1} Continual learning case study. From left to right:
(a) carbon footprints $\mathrm{C}^{(t_{0})}(n_{0})$ and corresponding
number of FL rounds ($n_{0}$) during initial training stage $t_{0}$,
for varying population of active learners $K_{a}=1\sim9$ and $K=9$
radars; (b) carbon footprints $\mathrm{C}^{(t_{i})}(n_{i})$ and corresponding
rounds $(n_{i})$ averaged over $2$ subsequent re-training stages
($i=1,2$). For all cases, footprints are evaluated for CL (dashed
line), CFA (cross markers), and FA-D (diamond markers) as well as
$4$ communication efficiency profiles that comply with LTE, NB-IoT,
IEEE 802.15.4e and WiFi IEEE 802.11.ac implementations. Rounds $(n_{i})$
and learners $(K_{a})$ are compared with bound (\ref{eq:size}) obtained
using the parameters of Tab. \ref{parameters}.}
\vspace{-0.3cm}
\end{figure}

\textbf{Selection of the device population} ($K_{a}$). The example
in Fig. \ref{comp1} uses the requirements in Table \ref{Region_definitions}
as a guideline for the selection of the smallest population size $K_{a}$,
so that FL emits lower carbon than CL during initial ($i=0$) and
re-training ($i>0$) stages. The figure highlights the carbon footprints
and the number of rounds for both stages. In particular, Fig. \ref{comp1}(a)
highlights the carbon footprints $\mathrm{C}^{(t_{0})}$ and the number
of learning rounds $n_{0}$ that are measured during the initial training
phase ($t_{0}$) for varying number of active learners $K_{a}$. Fig.
\ref{comp1}(b) shows the same results, namely $\mathrm{C}^{(t_{i})}$
and $n_{i}$, averaged over the subsequent re-training stages, $i=1,2$.
The target loss is here $\overline{\xi}=0.1$ for all cases. 

Considering the initial training first, FA-D and CFA implemented over
small population sets, i.e., $K_{a}\leq3$, need a large number of
rounds ($n>150$) and training time. Bound (\ref{eq:size}) for region
$\mathcal{R}_{b(\mathbf{W})}$ gives some practical guidelines for
optimizing $K_{a}$ population. For $\frac{b(\mathbf{W})}{b(\mathcal{E}_{k})}\simeq0.035$,
since $b[\mathcal{E}_{k}(t_{0})]=31$ MB, and the other parameters
in Tab. \ref{parameters}, the requirement (\ref{eq:size}) becomes
$K_{a}<\tfrac{257}{n}$. Considering CFA, the minimum population size
that satisfies (\ref{eq:size}) is $K_{a}=6$ since $n_{\mathrm{CFA}}=40$
rounds. Larger populations, i.e., $K_{a}\geq8$, increase the cost
per round, with no significant savings in terms of training time.
$K_{a}=4$ can be chosen for FA-D as $n_{\mathrm{FA-D}}=40$ rounds.
Using (\ref{eq:downlink}) for $K_{a}=4$, the required DL efficiency
should comply with $\mathrm{\tfrac{EE_{D}}{EE_{U}}>6.2}$: on the
other hand, since condition on computing cost (\ref{eq:training})
is satisfied for $\mathrm{CI}_{k}=0.97$ and $\varphi_{k}=0.36$,
as $K_{a}\cdot\tfrac{0.36}{0.95}<1.67$, a lower DL efficiency can
be tolerated.

Focusing now on model re-training stage, FA-D and CFA leave a smaller
footprint than CL for all choices of $K_{a}$ as they need few rounds
($<20$) for model update. With $\frac{b(\mathbf{W})}{b(\mathcal{E}_{k})}\simeq0.057$,
since now $b[\mathcal{E}_{k}(t_{i})]=19$ MB, the requirement (\ref{eq:size})
for FA-D becomes $K_{a}<\tfrac{158}{n}$ and it is satisfied for $K_{a}=1$
as $n_{\mathrm{FA-D}}<12$. For CFA, it is $K_{a}=3$ since $n_{\mathrm{CFA}}<20$.
For example, FA-D with only $K_{a}=1$ learner per round emits $1.1$
gCO2-eq per re-training stage (over LTE networks), which corresponds
to a carbon emission of $0.4$ kgCO2-eq per year, assuming $1$ re-training/day.
For the same setup, CL would cost $5.4$ gCO2-eq, corresponding to
roughly to $2$ kgCO2-eq per year.

To sum up, the results suggest that both FA-D and CFA are sustainable
choices for re-training, as they avoid unnecessary data uploads. On
the other hand, for initial training, CL and FL are both competitive
and should be considered carefully based on the available communication
interface and sustainability conditions (\ref{eq:downlink})-(\ref{eq:training}).
Notice that the availability of low-power sidelink communications
makes CFA the preferred choice. 

\subsection{Reinforcement learning for robot motion planning\label{subsec:Reinforcement-learning}}

According to the scenario of Fig. \ref{table_usecases}(b) and the
parameters of Tab. \ref{parameters}, the considered RL setting features
$K=5$ networked robots that collaboratively learn an optimized sequence
of motions, to follow an assigned trajectory, highlighted in green.
Since the goal is to quantify the carbon footprint of the distributed
learning processes against CL in a realistic setup, the motion control
problem is simplified by letting the robots move on a 2D regular grid
space consisting of $40$ landmark points, while the action space
consists of $4$ motions: Forward (F), Backward (B), Left (L), and
Right (R). Each robot explores a different site area collecting new
training data in real-time: data are obtained from two cameras, namely
a standard RGB camera and a short-range Time Of Flight (TOF) one \cite{terabee}.
Exploration of the environment and training data collection is responsible
for additional energy consumption (quantified here as $6.6$ Wh) that
depends on robot/servos hardware. 

RL and DeepMind model \cite{rl} are used to train a policy that maps
observations of the workplace to a set of actions, namely robot motions,
while trying to maximize a long-term reward\footnote{Devices get a larger reward whenever they approach the desired trajectory:
code, data and position-reward lookup table are described in the repository:
https://github.com/labRadioVision/Federated-DQL. Accessed: Mar. 2022.}. Exploration and model exploitation follow the $\varepsilon$-greedy
method while, in this example, exploration takes $20$ robot motions
for each learning round. DeepMind model consists of $5$ trainable
layers and $1.3$M parameters: model footprint from MQTT payload is
$b(\mathbf{W})=5.6$ MB. Notice that for CL the training data is moved
on every round, therefore $\alpha=n$. CFA and FA are considered here
with the setup highlighted in Sect. \ref{subsec:Adaptations-to-continual,},
with $K_{a}=K$. Data footprint is $b(\mathcal{E}_{k})=24.6$ MB,
therefore $\frac{b(\mathbf{W})}{b(\mathcal{E}_{k})}\simeq0.23$: with
such setup, condition (\ref{eq:size}) for RL is satisfied and requirement
for region $\mathcal{R}_{b(\mathbf{W})}$ becomes $\frac{b(\mathbf{W})}{b(\mathcal{E}_{k})}<\frac{K}{K_{a}}$.

\begin{figure}[!t]
\center\includegraphics[scale=0.58]{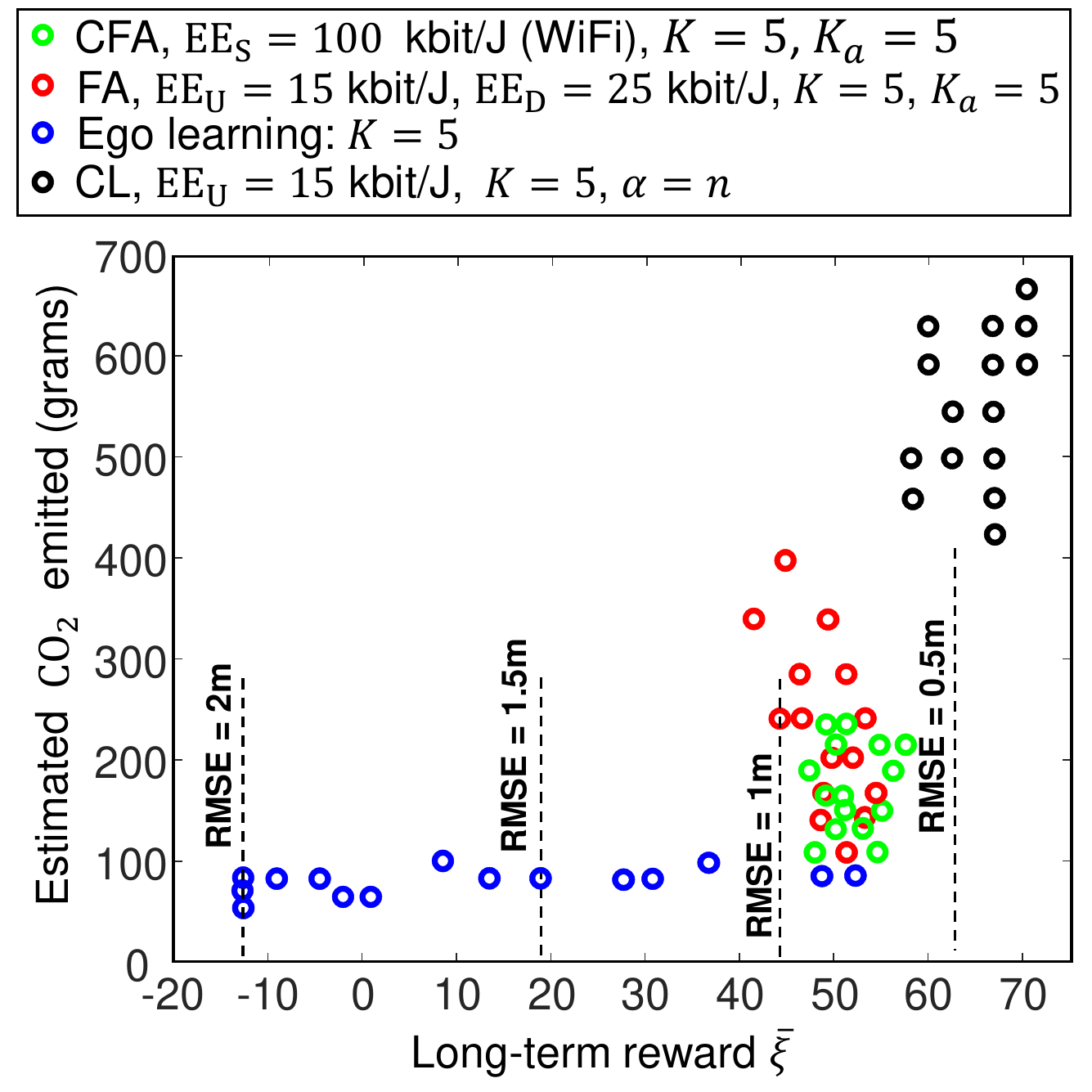} 
 \protect\caption{\label{RL} Reinforcement Learning for robot navigation: estimated
CO2 emissions for each policy optimization vs the running reward.
Each reward value corresponds to an average trajectory error compared
with the desired one, measured by root mean squared error (RMSE).
Specific parameters are in Tab. \ref{parameters}. CL is in black,
FA in red, CFA in green and ego learning in blue. Each point of the
scatter plot represents one simulation episode.}
\end{figure}
\textbf{Impact of target reward/accuracy.} Fig. \ref{RL} shows the
estimated carbon emissions versus the obtained reward considering
CL (black), FA (red), CFA (green) and ego learning (blue) policies.
For the opportunistic/ego learning approach \cite{commmag}, the robots
disable the radio interface and train their local models using the
training data from on-board sensors only. Each point in the scatter
plot of Fig. \ref{RL} corresponds to the carbon footprint measured
in one simulation episode. On each new episode, the observed footprint
might vary due to the random exploration phase. In ego learning, the
robots explore the environment for a longer time, thus training takes
around $25$ h corresponding to $4500\sim5000$ rounds since servos
need $1$ s per motion. For comparison, CL takes $90$ minutes on
average ($250\sim300$ rounds), while FA needs $180$ minutes ($500\sim550$
rounds) and CFA $150$ minutes ($400\div450$ rounds).

RL generally requires more rounds than continual learning to converge,
and cause large emissions when considering the whole training process.
Ego learning gives the lowest footprint as consumption is only due
to the robot and the LO. However, it experiences lower rewards, from
$-10$ to $30$ as the learned trajectory is far from the desired
one, with corresponding Root Mean Squared Errors\footnote{Positions far from the desired trajectory have small rewards: the
smaller the reward the larger the RMSE.} (RMSE) between $1.5$ m and $2$ m. CL converges faster at the cost
of a large footprint, about $2$ times larger than FA/CFA and $6$
times larger than ego learning. However, rewards are close to $70$,
that corresponds to learned trajectories with RMSE $0.5$ m. Compared
with CL, FA and CFA performance are now close since all robots are
active; energy savings are obtained at the cost of rewards penalties,
scaling down to $40\sim60$, with RMSE between $0.5$ m and $1$ m.
Considering ego and CL as extreme cases, where the former gives low-accuracy
and small footprint, while the latter high-accuracy and large footprint,
FL is a promising tradeoff solution, as trading accuracy (reward)
with environmental footprint.

\section{Conclusions\label{sec:Conclusions-and-open}}

The article proposed a novel framework for the analysis of energy
and carbon footprints in distributed and federated learning. A tradeoff
analysis was characterized for vanilla FL and decentralized, consensus-driven
learning, compared with centralized training. The analyzed algorithms
are suitable for low-power embedded wireless devices with constrained
memory, typically adopted in industrial IoT processes requiring low
latency. The paper also explored operational conditions applicable
to both conventional FL methods and generalized distributed learning
policies targeting carbon-efficiency. Minimal requirements for FL
are obtained analytically and have been applied in different scenarios
to steer the choice towards sustainable designs. 

Carbon equivalent emissions have been analyzed for two industry relevant
5G verticals, using real datasets and a test-bed characterized by
physical IoT devices implementing real-time model training on top
of the MQTT transport, tailored to support various learning processes.
For each considered case, centralized versus distributed training
impact is discussed for continual and reinforcement learning problems.
Sustainability of the FL depends in many cases on model vs data footprints,
as well as the population of active devices, that must be selected
to balance energy budget and training data quality. Downlink vs uplink
(or sidelink) communication efficiency and the amount of available
green energy are additional key factors. The proposed CFA and FA with
deep sleep mode (FA-D) methods have been shown to be effective as
they minimize the number of active learners at each learning round.
The selection of the learner population size according to the proposed
design patterns provides significant savings per round ($20\%\sim30\%$
and higher) compared with baselines. CFA has an advantage over FA-D
provided that an efficient sidelink communication interface is available:
the co-design of learning and communication is thus of high importance.
Finally, the energy footprints of reinforcement learning need much
more careful considerations than conventional supervised learning,
due to the considerable carbon emissions. Results indicate that FL
is a promising solution as it trades off energy and accuracy compared
with the extreme cases of ego (low rewards/footprint) and centralized
training (high rewards/footprint).

\section*{Appendix\label{sec:Appendix}}

\subsection*{Communication costs: FA and FA-D design constraints}

Using the carbon footprints shown in Tab. \ref{carbon_foots}, the
constraint $\mathrm{C}_{\mathrm{FA}}^{(\mathrm{C})}<\mathrm{C}_{\mathrm{CL}}^{(\mathrm{C})}$
in region $\mathcal{R}_{\mathrm{DU}}$ (\ref{eq:regiondl}) implies
that, after straightforward manipulation, the following condition
has to be met
\begin{equation}
\mathrm{\frac{EE_{D}}{\mathrm{EE}_{\mathrm{U}}}\cdot}\left[\mathcal{\mathcal{H}}\left(\frac{b(\mathbf{W})}{b(\mathcal{E})}\right)-\frac{1}{K_{a}}\sum_{k=1}^{K_{a}}\mathrm{CI}_{k}\right]>\gamma\cdot\frac{K}{K_{a}}\mathrm{CI}_{0}\label{eq:e}
\end{equation}
with 
\begin{equation}
\mathcal{\mathcal{H}}\left(\frac{b(\mathbf{W})}{b(\mathcal{E})}\right)=\frac{\alpha}{n\cdot K_{a}}\frac{b(\mathcal{E})}{b(\mathbf{W})}\sum_{k=1}^{K}\sigma_{k}\mathrm{CI}_{k}\label{eq:H}
\end{equation}
where $\sigma_{k}$, defined in (\ref{eq:farho}) with $\sum_{k=1}^{K}\sigma_{k}=1,$
counts the number of the local ($k$) examples out of the number of
the global ones, respectively. Eq. (\ref{eq:downlink}) is then obtained
by setting $\mathrm{CI}_{k}=\mathrm{CI}$, $\forall k$. Considering
FA-D, the equation (\ref{eq:e}) becomes $\mathrm{\frac{EE_{D}}{\mathrm{EE}_{\mathrm{U}}}\cdot}\mathcal{\mathcal{H}}\left(\frac{b(\mathbf{W})}{b(\mathcal{E})}\right)>\gamma\mathrm{CI}_{0}$. 

For $\mathrm{\frac{EE_{D}}{\mathrm{EE}_{\mathrm{U}}}}$ large enough
as assumed for derivation of region $\mathcal{R}_{b(\mathbf{W})}$
in problem (\ref{eq:regionbw}), the condition $\max[\mathrm{C}_{\mathrm{FA}}^{(\mathrm{C})},\mathrm{C}_{\mathrm{FA-D}}^{(\mathrm{C})}]<\mathrm{C}_{\mathrm{CL}}^{(\mathrm{C})}$
is guaranteed as long as 
\begin{equation}
\mathcal{\mathcal{H}}\left(\frac{b(\mathbf{W})}{b(\mathcal{E})}\right)>\frac{1}{K_{a}}\sum_{k=1}^{K_{a}}\mathrm{CI}_{k},\label{eq:11}
\end{equation}
that corresponds to 
\begin{equation}
\frac{b(\mathbf{W})}{b(\mathcal{E})}\cdot K<\frac{\alpha}{n}\cdot\frac{\sum_{k=1}^{K}\sigma_{k}\mathrm{CI}_{k}}{\sum_{k=1}^{K_{a}}\mathrm{CI}_{k}}.\label{eq:ee}
\end{equation}
Constraint (\ref{eq:ee}) can be further simplified by letting, $\forall k,$
$\sigma_{k}\simeq1/K,$ namely $b(\mathcal{E}_{k})\simeq\frac{b(\mathcal{E})}{K}$,
and $\mathrm{CI}_{k}=\mathrm{CI}$, to obtain (\ref{eq:size}) with
$\Gamma=1$.

\subsection*{Communication costs: CFA design constraints}

Using Tab. \ref{carbon_foots} and (\ref{eq:H}), the constraint $\mathrm{C}_{\mathrm{CFA}}^{(\mathrm{C})}<\mathrm{C}_{\mathrm{CL}}^{(\mathrm{C})}$
for region (\ref{eq:regionsl}) results in 
\begin{equation}
\frac{\mathrm{EE_{S}}}{\mathrm{EE}_{\mathrm{U}}}\cdot\mathcal{\mathcal{H}}\left(\frac{b(\mathbf{W})}{b(\mathcal{E})}\right)>\frac{N}{K_{a}}\sum_{k=1}^{K_{a}}\mathrm{CI}_{k}\label{ee2}
\end{equation}
that simplifies to (\ref{eq:EES}) when, $\forall k$, it is $b(\mathcal{E}_{k})\simeq\frac{b(\mathcal{E})}{K}$,
and $\mathrm{CI}_{k}=\mathrm{CI}$. Furthermore, 
\begin{equation}
\mathrm{\frac{EE_{S}}{\mathrm{EE_{U}}}}+\gamma\cdot K\cdot\frac{\mathrm{CI}_{0}}{\sum_{k=1}^{K_{a}}\mathrm{CI}_{k}}\cdot\mathrm{\frac{EE_{S}}{\mathrm{EE_{D}}}}>N\label{eq:energy}
\end{equation}
guarantees lower carbon footprint \emph{per round} than FA, namely
$\frac{\mathrm{C}_{\mathrm{CFA}}^{(\mathrm{C})}}{n_{\mathrm{CFA}}}<\frac{\mathrm{C}_{\mathrm{FA}}^{(\mathrm{C})}}{n_{\mathrm{FA}}}.$
The comparison of CFA vs FA-D gives the same result shown in (\ref{eq:energy})
but a more stringent condition as $K$ must be replaced with $K_{a}$.
As expected, considering an individual round, CFA footprint is closer
to FA-D as both strategies target the minimization of the active device
population.

In WWAN, or cellular settings, D2D connectivity is replaced by UL
and DL communications, namely $[\mathrm{EE}_{\mathrm{S}}]^{-1}=[\mathrm{EE}_{\mathrm{D}}]^{-1}+[\mathrm{EE}_{\mathrm{U}}]^{-1}$.
In this case the constraint (\ref{ee2}) becomes
\begin{equation}
\frac{\mathrm{EE_{D}}}{\mathrm{EE}_{\mathrm{U}}}\cdot\left[\mathcal{\mathcal{H}}\left(\frac{b(\mathbf{W})}{b(\mathcal{E})}\right)-\frac{N}{K_{a}}\sum_{k=1}^{K_{a}}\mathrm{CI}_{k}\right]>\frac{N}{K_{a}}\cdot\sum_{k=1}^{K_{a}}\mathrm{CI}_{k},
\end{equation}
while for $\mathrm{\frac{EE_{D}}{\mathrm{EE}_{\mathrm{U}}}}$ large
enough, sustainability is guaranteed as long as 
\begin{equation}
\mathcal{\mathcal{H}}\left(\frac{b(\mathbf{W})}{b(\mathcal{E})}\right)>\frac{N}{K_{a}}\sum_{k=1}^{K_{a}}\mathrm{CI}_{k}.\label{eq:12}
\end{equation}
Comparing (\ref{eq:11}) with (\ref{eq:12}), the condition $\max[\mathrm{C}_{\mathrm{CFA}}^{(\mathrm{C})},\mathrm{C}_{\mathrm{FA}}^{(\mathrm{C})},\mathrm{C}_{\mathrm{FA-D}}^{(\mathrm{C})}]<\mathrm{C}_{\mathrm{CL}}^{(\mathrm{C})}$
for the derivation of region $\mathcal{R}_{b(\mathbf{W})}$ in (\ref{eq:regionbw}),
can be rewritten as in (\ref{eq:12}). Finally, replacing $\forall k$
$\mathrm{CI}_{k}=\mathrm{CI}$, we obtain (\ref{eq:size}) now with
$\Gamma=N$.

\end{document}